\documentclass[runningheads]{llncs}

\usepackage[preprint]{eccv}

\usepackage{eccvabbrv}

\usepackage{graphicx}
\usepackage{booktabs}
\usepackage{multirow}
\usepackage{colortbl} % 必须：用于单元格颜色
\usepackage{xcolor}   % 必须：颜色支持
\usepackage{array}    % 推荐：更好的列控制
\usepackage{algorithm} 
\usepackage{algorithmic}
\usepackage{listings}
\usepackage{makecell}
\usepackage{pgfplots}
\usepackage{tcolorbox}
\usepackage{calc}
\definecolor{codegreen}{rgb}{0,0.6,0}
\definecolor{codegray}{rgb}{0.5,0.5,0.5}
\definecolor{codepurple}{rgb}{0.58,0,0.82}
\definecolor{backcolour}{rgb}{0.95,0.95,0.92}
\definecolor{item}{RGB}{15, 86, 157}
\definecolor{query}{RGB}{169, 16, 3}
\definecolor{lightblue}{rgb}{0.6, 0.8, 0.9}
\definecolor{darkblue}{rgb}{0.2,0.4,0.6}
\definecolor{darkgreen}{rgb}{0, 0.55, 0.12}
\definecolor{darkred}{rgb}{0.6,0,0}
\definecolor{-}{rgb}{0.25,0.41,0.88}
\definecolor{+}{rgb}{0.70,0.13,0.13}

\definecolor{table_color}{RGB}{239,246,251}
\definecolor{good}{RGB}{58,113,104}
\definecolor{bad}{RGB}{180,0,0}
\definecolor{malboxborder}{RGB}{180,0,0}
\definecolor{benboxborder}{RGB}{81,91,131}
\definecolor{malboxbg}{RGB}{255,250,250}
\definecolor{benboxbg}{RGB}{251,252,254}
\definecolor{titlebg}{RGB}{77,77,77}
\definecolor{rtnbg}{RGB}{245, 245, 245}       % 浅灰色背景给 RTN
\definecolor{gptqbg}{RGB}{230, 240, 255}      % 浅蓝色背景给 GPTQ (或者用浅橙色区分)
\definecolor{besthighlight}{RGB}{255, 235, 205} % 最优值高亮色
\newtcolorbox{promptbox}{
    colback=white,
    colframe=black!70,
    boxrule=0.8pt,
    arc=4pt,
    left=6pt, right=6pt, top=6pt, bottom=6pt
}

\newtcolorbox{baselinebox}{
    colback=red!5, % 浅红背景暗示错误
    colframe=bad,
    boxrule=0.5pt,
    arc=3pt,
    left=4pt, right=4pt, top=4pt, bottom=4pt,
    before skip=2mm, after skip=2mm
}

\newtcolorbox{freequantbox}{
    colback=green!5, % 浅绿背景暗示正确
    colframe=darkgreen,
    boxrule=0.5pt,
    arc=3pt,
    left=4pt, right=4pt, top=4pt, bottom=4pt,
    before skip=2mm, after skip=2mm
}
\usepackage[accsupp]{axessibility}  % Improves PDF readability for those with disabilities.

\usepackage{hyperref}

\usepackage{orcidlink}

\begin{document}

\title{BATQuant: Outlier-resilient MXFP4 Quantization via Learnable Block-wise Optimization} 

\titlerunning{BATQuant: Outlier-resilient MXFP4 Quantization}

\author{Ji-Fu Li\inst{1} \and
Manyi Zhang\inst{1}\thanks{Corresponding author. \\ \textit{Preprint.}} \and
Xiaobo Xia\inst{2} \and Han Bao\inst{1} \and Haoli Bai\inst{1} \and Zhenhua~Dong\inst{1} \and Xianzhi Yu\inst{1} }

\authorrunning{Li et al.}

\institute{
Huawei Technologies \\
\and
University of Science and Technology of China \\
\email{\{lijifu4, zhangmanyi6\}@huawei.com}
}
\maketitle
\begin{abstract}
  Microscaling floating-point (MXFP) formats have emerged as a promising standard for deploying Multi-modal Large Language Models (MLLMs) and Large Language Models (LLMs) on modern accelerator architectures. However, existing Post-Training Quantization (PTQ) methods, particularly rotation-based techniques designed for integer formats, suffer from severe performance collapse when applied to MXFP4. Recent studies attribute this failure to a fundamental format mismatch: global orthogonal rotations inadvertently transfer outlier energy across quantization blocks, inducing new outliers that disrupt local block-wise scaling, while often creating bimodal activation distributions that underutilize the limited quantization range. 
  To address these issues, we propose BATQuant (\textbf{B}lock-wise \textbf{A}ffine \textbf{T}ransformation), which restricts transformations to align with MXFP granularity to prevent cross-block outlier propagation, while relaxing orthogonality constraints to optimize distribution shaping. To ensure parameter efficiency, we introduce \textbf{G}lobal and \textbf{P}rivate \textbf{K}ronecker (GPK) decomposition to effectively reduces storage and runtime overhead and incorporate Block-wise Learnable Clipping to suppress residual outliers. Extensive experiments on both MLLMs and LLMs demonstrate that BATQuant establishes new state-of-the-art results under aggressive W4A4KV16 configurations, recovering up to \textbf{96.43\%} of full-precision performance on multimodal benchmarks and clearly outperforming existing methods across diverse tasks.
\end{abstract}

\section{Introduction}
\label{sec:intro}
Multi-modal Large Language Models (MLLMs) and Large Language Models (LLMs) have recently revolutionized artificial intelligence, demonstrating remarkable capabilities in bridging visual perception with linguistic reasoning~\cite{liu2023visual,bai2025qwen3, wang2025internvl3, zeng2025glm, hong2025glm, team2025kimi, wu2024deepseek,luo2025next,luo2025gui}. From autonomous driving to medical image analysis, these models are increasingly deployed in real-world scenarios where low latency and memory efficiency are paramount~\cite{liu2025quantization, zhu2024survey, xu2024survey, wang2024model,liu2025mtp}. However, the ever-growing scale of MLLMs and LLMs, often comprising billions of parameters, imposes prohibitive costs on memory bandwidth and computational resources, hindering their deployment on edge devices and resource-constrained platforms.

Post-Training Quantization (PTQ) has emerged as a key solution to mitigate these costs. While integer quantization has been widely studied, the recent emergence of microscaling floating-point formats (MXFP) offers a promising alternative~\cite{rouhani2023microscaling, agarwal2025gpt}. Supported by next-generation hardware~\cite{amd2025cdna4, choquette2023nvidia, tirumala2024nvidia}, MXFP4 utilizes block-wise scaling to better accommodate the long-tailed distributions inherent in activations, theoretically offering superior dynamic range compared to fixed-point formats. Despite this hardware readiness, achieving accurate 4-bit quantization for MLLMs under the MXFP format remains an unsolved challenge~\cite{zhang2026benchmarking, zhao2026unleashing}.

While existing state-of-the-art PTQ methods are predominantly designed for INT formats~\cite{shao2023omniquant,ma2024affinequant,li2024svdquant,wei2023outlier,wang2025bitnet, li2025mbq, qin2026veq,liu2026freeact}, their applicability to MXFP formats is contested. Specifically, popular rotation-based techniques (e.g., QuaRot~\cite{ashkboos2024quarot} and SpinQuant~\cite{liu2024spinquant}), which excel in INT4 by spreading outliers via orthogonal transformations, suffer from severe performance collapse when applied to MXFP4~\cite{egiazarian2025bridging, meng2026arcquant}. 
Recent studies~\cite{shao2025block, egiazarian2025bridging} have attributed this failure to the incompatibility between global rotations and the fine-grained quantization settings of MXFP, and further propose block-wise rotation transformation methods. However, these approaches still fail to mitigate extreme outliers within certain blocks, and the Hadamard transform further introduces a bimodal distribution problem (see Figure~\ref{fig:BRQ}).

To bridge this gap, in this paper, we introduce BATQuant. The core of our method is the \textit{Block-wise Affine Transformation}~(BAT). Unlike global rotations, BAT restricts the transformation scope to align strictly with the MXFP quantization granularity (e.g., 32 elements). This design prevents the cross-block energy transfer of outliers, ensuring that each block's scaling factor accurately captures its local dynamic range.  
Moreover, we relax the orthogonality constraint and learn the optimal affine matrices tailored to the MXFP format to minimize quantization error. To address the storage overhead caused by learnable block-wise affine transformations, we further introduce the \textit{Global and Private Kronecker} (GPK) decomposition that drastically reduces parameter counts by sharing a global transformation basis across blocks while retaining block-specific private components. Finally, we incorporate \textit{Block-wise Learnable Clipping}, which dynamically adapts thresholds to suppress residual outliers within quantization blocks.

\begin{figure}[t]
    \centering
    \includegraphics[width=1\linewidth]{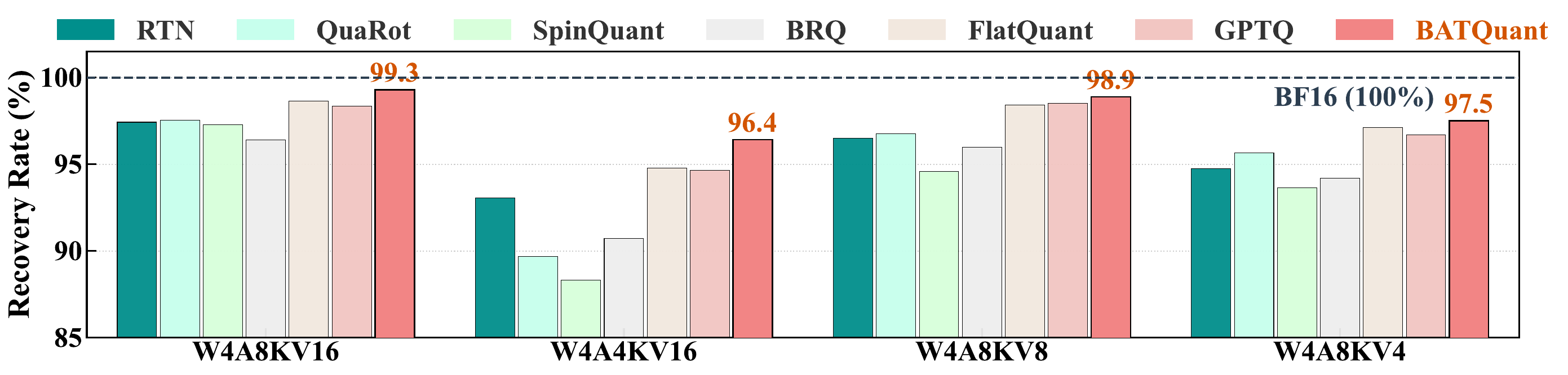}
    \caption{\textbf{Quantization performance on Qwen3-VL-8B-Instruct across various methods.}  Our method yields superior results compared to baselines across all bit-width settings. The advantage is particularly substantial in the W4A4 setting, where our method clearly outperforms existing methods.}
    \label{fig:mllm_recovery}
\end{figure}

We validate our BATQuant extensively on both MLLMs and LLMs. Our method achieves \textbf{near-lossless} performance on \texttt{W4A8KV16} with an accuracy recovery rate exceeding \textbf{99\%}. Furthermore, it establishes the new state-of-the-art results under aggressive \texttt{W4A4KV16} configurations, recovering up to \textbf{96.43\%} on multimodal benchmarks, significantly outperforming existing methods (see Figure~\ref{fig:mllm_recovery}). Our main contributions are summarized as follows:
\begin{itemize}

    \item We propose BATQuant, featuring a Block-wise Affine Transformation that aligns with MXFP granularity to prevent energy transfer across blocks and address the bimodal distribution problem for effective quantization. Additionally, we incorporate Global and Private Kronecker decomposition for parameter efficiency.

    \item We evaluate BATQuant on both MLLMs and LLMs, such as Qwen3-8B-VL-Instruct~\cite{bai2025qwen3} and Qwen3-8B~\cite{yang2025qwen3}, covering a wide range of challenging settings. The effectiveness is validated, ranging from knowledge understanding to complex reasoning benchmarks, setting new state-of-the-art results in most scenarios. 
\end{itemize}

\section{Preliminary}
\label{sec:preliminary}
\noindent\textbf{Microscaling Floating-Point Definition.} The MXFP, proposed by OCP~\cite{rouhani2023microscaling}, is a family of floating-point formats that employ block-wise quantization. An MXFP format is defined by three components: a sign bit ($S$), an exponent ($E$), and a mantissa ($M$). Each MXFP format uses a fixed block size of 32 elements, with all values in a block sharing a common scaling factor represented in UE8M0 format (8-bit exponent, no mantissa). The standard MXFP4 (E2M1) format uses 1 sign bit, 2 exponent bits, and 1 mantissa bit. This configuration represents 7 distinct positive values: \{0.5, 1.0, 1.5, 2.0, 3.0, 4.0, 6.0\}, along with their negatives and zero. MXFP8 offers two variants, E4M3 and E5M2. Here, we adopt E4M3 for MXFP8, as a larger mantissa width is more crucial for the performance of fine-grained quantization~\cite{mishra2025recipes, chen2025int}.

\vspace{5pt}
\noindent\textbf{Related Work.} Initial research on LLM quantization primarily explored integer-based formats~\cite{xiao2023smoothquant, yu2025mquant, frantar2022gptq, lin2024duquant, lin2024awq, hu2025moequant}. As NVFP and MXFP formats gain hardware support, quantization accuracy under these formats is also drawing increasing attention~\cite{hu2026m2xfp, lee2025mx+, liu2025micromix, zhang2025sageattention3, xin2026quantization}. Prior work has shown that MXFP8 achieves lossless quantization, whereas MXFP4 suffers from significant accuracy degradation~\cite{zhang2026benchmarking}. For the low-bit scenarios, e.g., 4-bit quantization, outliers are considered as a severe impediment. The primary methods for suppressing outliers include rotation transformations~\cite{tseng2024quip, li2025mbq, huang2024rolora} and affine transformations~\cite{sun2024flatquant, ma2024affinequant}. The rotation-based methods, such as QuaRot~\cite{ashkboos2024quarot} and SpinQuant~\cite{liu2024spinquant}, unlike their success in INT4 quantization, underperform even basic RTN when applied to MXFP4. Such global rotations mix dimensional information, suppressing outliers and kurtosis, thus disrupting the local statistical properties of fine-grained formats. To address the incompatibility between rotation-based techniques and MXFP4, BRQ~\cite{shao2025block} is proposed to utilize block-wise rotation quantization to mitigate outliers and prevent amplifying small-value blocks. MR-GPTQ~\cite{egiazarian2025bridging}, a GPTQ variant optimized for FP4, similarly employs block-wise Hadamard transforms and format-specific adjustments to accommodate FP4's unique properties. Affine transformation-based methods, such as FlatQuant~\cite{sun2024flatquant}, overcome the energy-conservation constraints inherent in rotation-based transformations and enhance quantization accuracy by employing affine transformations. Nevertheless, previous methods still suffer from significant accuracy degradation on MXFP4 quantization, particularly on complex reasoning tasks~\cite{zhang2026benchmarking}.

\begin{figure}[!t]
    \centering
    \begin{subfigure}[b]{0.48\linewidth}
        \includegraphics[width=\linewidth]{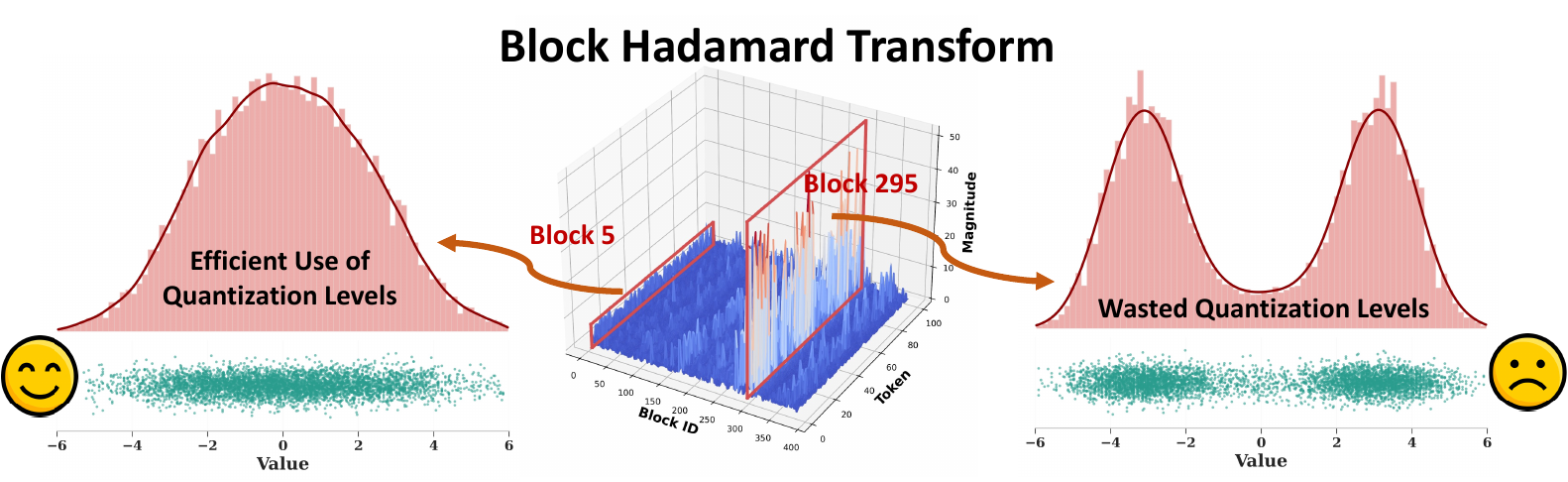}
        \caption{BRQ}
        \label{fig:BRQ}
    \end{subfigure}
    \hfill
    \begin{subfigure}[b]{0.48\linewidth}
        \includegraphics[width=\linewidth]{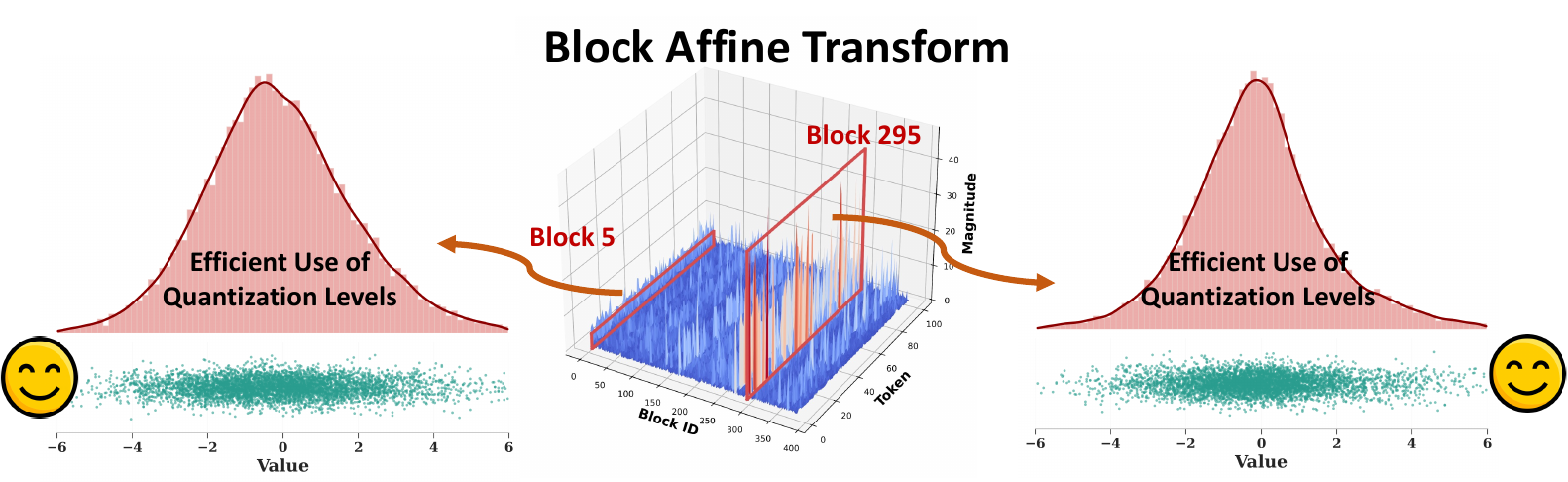}
        \caption{BATQuant}
        \label{fig:BAT}
    \end{subfigure}
    \caption{Activation distributions for the \texttt{down\_proj} module in layer 35 of Qwen3-8B. The central 3D plots illustrate the activations after transformation. We specifically extract Block 5 (without outliers) and Block 295 (with extreme outliers), and visualize the values after scaling factor division but prior to rounding. (a) After applying the block Hadamard transform, block 295 exhibits a bimodal distribution, leading to inefficient utilization of the bit width. (b) After the block affine transformation, block 295 shows reduced magnitude compared to subplot (a) while effectively leveraging the floating-point quantization grids.}
    \label{fig:BRQ_BAT_vis}
\end{figure}

\vspace{5pt}
\noindent\textbf{Observations and Motivation.} We find that block-wise rotation still struggles to suppress extreme outliers in specific blocks, and the Hadamard transform further introduces a bimodal distribution problem. Specifically, we visualize the activations after block-wise Hadamard transformation on Qwen3-8B in Figure~\ref{fig:BRQ}. We observe that although the block-wise Hadamard transform reduces the magnitude for the vast majority of blocks, since Hadamard matrices are composed of $\{+1, -1\}$ values, certain blocks with extreme outliers exhibit a bimodal distribution. This results in wasted bit-width and introduces larger quantization errors~\cite{cook2025four}. Therefore, to address these challenges, we propose BATQuant. As shown in Figure~\ref{fig:BAT}, BATQuant effectively alleviates outliers, while ensuring that the post-transformation data distribution remains amenable to floating-point quantization.

\begin{figure}[!t]
    \centering
    \includegraphics[width=1.0\linewidth]{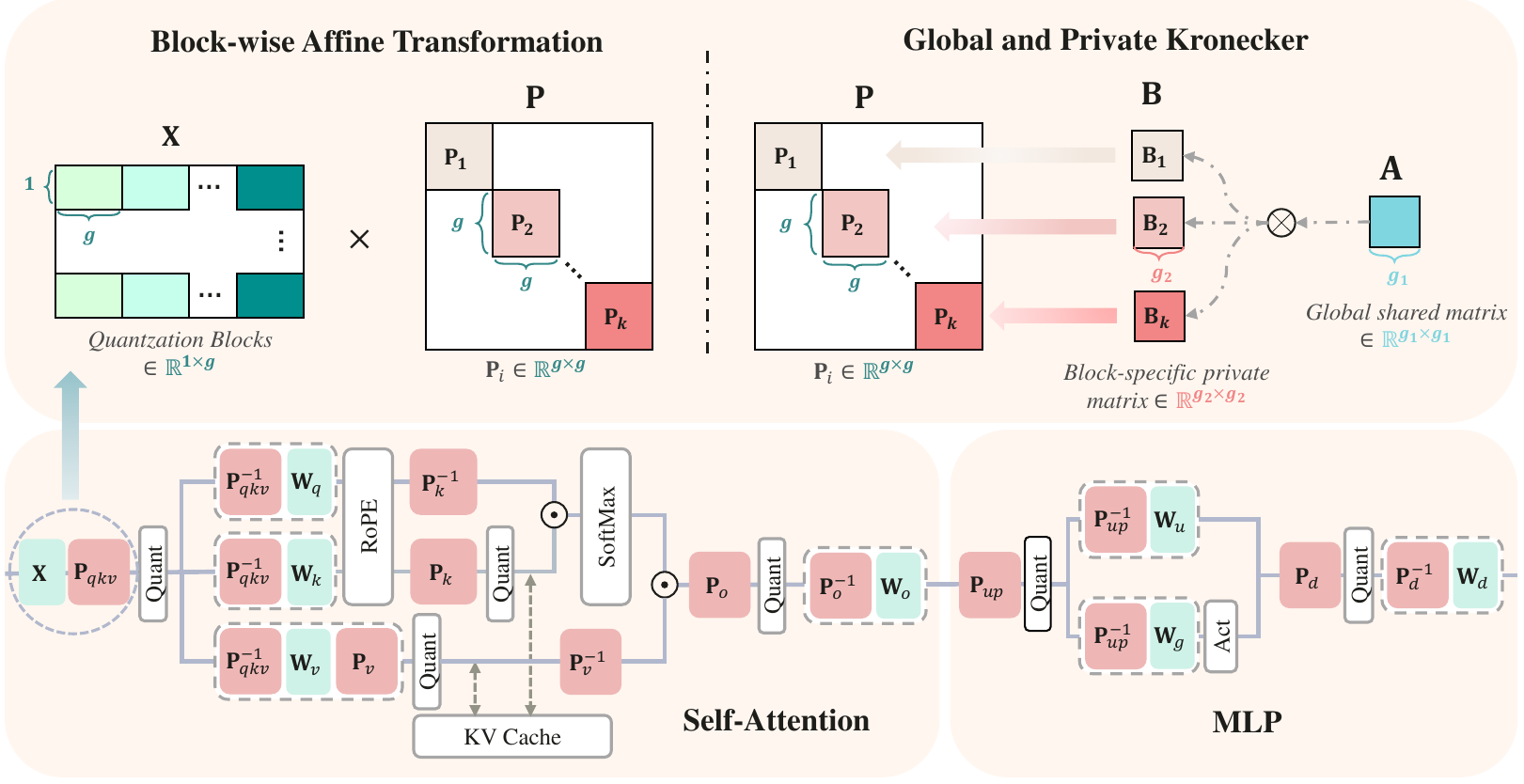}
    \caption{\textbf{The overall framework of BATQuant.} \textit{Bottom:} Integration of BATQuant into the Transformer architecture. Weight-side transformations are fused offline into the linear layers, while activation-side transformations are applied online. \textit{Top:} Exemplary view of the \textit{Block-wise Affine Transformation}, where inputs are partitioned into MXFP-aligned blocks. Each block transformation is decomposed via the \textit{Global and Private Kronecker}.} 
    \label{fig:overview}
\end{figure}
\vspace{10pt}
\section{Method}
\label{sec:method}
In this section, we present BATQuant with the framework illustrated in Figure~\ref{fig:overview}. We first introduce learning optimal block-wise affine transformations in Section~\ref{sec:affine_formulation}. Afterward, we discuss its integration with the Transformer architecture in Section~\ref{sec:integration}. Note that we provide a detailed algorithm flow of our BATQuant in Appendix \ref{app:algorithm}.

\vspace{10pt}
\subsection{Block-wise Affine Transformation}
\label{sec:affine_formulation}
Consider a standard linear layer computation $\mathbf{Y}=\mathbf{X}\mathbf{W}^{\top}$, where $\mathbf{X} \in \mathbb{R}^{S\times N} $ represents activations and $\mathbf{W} \in \mathbb{R}^{M \times N}$ denotes weights. The primary objective is to find the best affine transformation $\mathbf{P}^{\star} \in \mathbb{R}^{N \times N}$ for each linear layer to quantize:
$$
\mathbf{P}^{\star} = \mathop{\arg\min}_{\mathbf{P}} ||\mathbf{Y}-\mathcal{Q}(\mathbf{X}\mathbf{P})\mathcal{Q}(\mathbf{P}^{-1}\mathbf{W}^{\top})||^2_{F}.
$$
Instead of learning a single global matrix, we partition the transformation matrix into $k$ disjoint blocks aligned with the MXFP quantization granularity $g$ (e.g., $g=32$). We then construct a block-diagonal affine matrix:
\begin{equation}
\mathbf{P} = \text{diag}(\mathbf{P}_1, \mathbf{P}_2, \dots, \mathbf{P}_k), \quad \text{where } \mathbf{P}_i \in \mathbb{R}^{g \times g}, N=k \cdot g.
\end{equation}
Here, each $\mathbf{P}_i$ is an independent and learnable affine transformation applied solely within the $i$-th quantization block. By restricting the transformation scope to the size of the MXFP block, our method ensures that outlier redistribution occurs only locally. This preserves the statistical independence of each quantization block, allowing the MXFP scaling factors to accurately capture the dynamic range of each block without interference from outliers of other blocks. 
\subsubsection{Global and Private Kronecker.}
Although the block-diagonal structure of $\mathbf{P}$ introduces inherent sparsity, the total number of learnable parameters remains $N \cdot g$. For large-scale models, storing such a matrix for every layer still incurs a significant memory cost. A straightforward approach to mitigate this is to apply Kronecker product decomposition to each $\mathbf{P}_i$, factorizing it into two smaller matrices $\mathbf{B}_i \otimes \mathbf{A}_i$, where $\mathbf{A}_i \in \mathbb{R}^{g_1 \times g_1}, \mathbf{B}_i \in \mathbb{R}^{g_2 \times g_2}$. The $g_1$ and $g_2$ respectively denote the size of $\mathbf{A}_i$ and $\mathbf{B}_i$ and we have MXFP quantization granularity $g=g_1 \cdot g_2$. We refer to this as \textit{Naive Kronecker}. However, since the block size $g$ is typically small (e.g., 32 in MXFP formats), the reduction in parameter count is marginal.

To address this limitation, we propose \textit{Global and Private Kronecker} (GPK). GPK decomposes each $\mathbf{P}_i$ into the product of a \textit{global shared matrix} $\mathbf{A}$ and a \textit{block-specific private matrix} $\mathbf{B}_i$:
\begin{equation}
    \mathbf{P}_i = \mathbf{B}_i \otimes \mathbf{A}, \quad \forall i \in \{1, \dots, k\},
\end{equation}
where $\mathbf{A}$ is shared across all $k$ blocks and $\mathbf{B}_i$ is unique to the $i$-th block. This design drastically reduces the storage requirement from $k \cdot (g_1^2+g_2^2)$ to $g_1^2+k \cdot g_2^2$. As shown in Table~\ref{tab:param_ana}, GPK significantly reduces the storage overhead, reducing the parameter count by more than 74\% and 79\% compared to FlatQuant and Naive Kronecker. Additionally, by leveraging the vectorization trick of the
Kronecker product, i.e., $\text{vec}(\mathbf{V})(\mathbf{B}_i \otimes \mathbf{A}) = \text{vec}(\mathbf{B}_i^{\top}\mathbf{V}\mathbf{A})$
for some $\mathbf{V} \in \mathbb{R}^{g_2\times g_1}$, GPK maintains efficient inference by preserving the low matrix multiplication complexity. Here, we provide the PyTorch-style pseudo code of the forward pass with GPK in Appendix \ref{app:algorithm}.
\begin{table}[t]
\centering
\caption{Comparison of decomposition methods on parameter counts and computational cost. For the example parameter count, we set the hidden dim $N=4096$ and the MXFP quantization granularity $g=32$. The size of decomposed matrix $\mathbf{A}_i$ and $\mathbf{B}_i$ are set to $g_1=8$ and $g_2=4$. The reported \textit{MatMul Complexity} refers to the computational cost of the activation transformation $\mathbf{X}\mathbf{P}$. }
\setlength{\tabcolsep}{4pt}
\label{tab:param_ana}
% \small % 如果表格太宽，可以使用 \small 或 \footnotesize
\resizebox{\linewidth}{!}{%
\begin{tabular}{l l l l c} % 列格式：左对齐，左对齐，居中，右对齐(数字)
\toprule
{Method} & {Decomposition} & MatMul Complexity &\# Params of $\mathbf{P}$ & {Example Count} \\ 
\midrule
FlatQuant       & Kronecker       &    $\mathcal{O}(SN^{\frac{3}{2}})$   & $2N$             & 8,192                    \\
\midrule
\multirow{3}{*}{{Ours}} 
                & w/o                & $\mathcal{O}(SNg)$   & $N \cdot g$      & 131,072                  \\
                & Naive Kronecker    & $\mathcal{O}(SN(g_1+g_2))$   & $k \cdot (g_1^2+g_2^2)$               & 10,240                   \\
                & \cellcolor{gray!15}{GPK}     &  \cellcolor{gray!15}$\mathcal{O}(SN(g_1+g_2))$    & \cellcolor{gray!15}$g_1^2+k \cdot g_2^2$               & \cellcolor{gray!15}{2,112}           \\
\bottomrule
\end{tabular}
}
% \vspace{-0.5em} % 微调表格下方的间距
\end{table}
\subsubsection{Block-wise Learnable Clipping.} While the block-wise affine transformation effectively smooths activation distributions, residual outliers may still persist within the quantization blocks, potentially dominating the quantization range of MXFP formats. To mitigate this, we introduce \textit{Block-wise Learnable Clipping}, a fine-grained strategy that adapts clipping thresholds to the local statistics of each quantization block. For the $i$-th block, the clipped values $\hat{\mathbf{x}}_i$ (and similarly for weights $\hat{\mathbf{w}}_i$) are computed as:
\begin{equation}
    \hat{\mathbf{x}}_i = \text{clip}\left(\mathbf{x}_i, \; \beta_i^{\text{min}}, \; \beta_i^{\text{max}} \right),
\end{equation}
where the dynamic bounds $\beta_i^{\text{min}}$ and $\beta_i^{\text{max}}$ are:
\begin{equation}
    \beta_i^{\text{min}} = \sigma(\alpha_i^{\text{min}}) \cdot \min(\mathbf{x}_i), \quad \beta_i^{\text{max}} = \sigma(\alpha_i^{\text{max}}) \cdot \max(\mathbf{x}_i).
\end{equation}
Here, $\min(\mathbf{x}_i)$ and $\max(\mathbf{x}_i)$ denote the minimum and maximum values within the $i$-th block, respectively, and $\sigma(\cdot)$ is the sigmoid function constraining the clipping ratios to $(0, 1)$. $\alpha_i$ is the learnable parameter specific to block $i$.
\subsubsection{The Training Objective.} Following previous work \cite{sun2024flatquant}, we optimize the block-wise affine transformations and clipping factors by minimizing the layer-wise quantization errors between the full-precision and quantized outputs over a small calibration set $\mathcal{D}_{\text{cal}}$:
\begin{equation}
    \Theta_l^* = \arg \min_{\Theta_l} \mathbb{E}_{\mathbf{X} \sim \mathcal{D}_{\text{cal}}} \left[ \left\| \mathcal{F}_l(\mathbf{X}) - \mathcal{\hat{F}}_l(\mathbf{X}; \Theta_l) \right\|_2^2 \right]
\end{equation}
where $\mathcal{F}_l(\cdot)$ and $\mathcal{\hat{F}}_l(\cdot)$ denote the full-precision layer $l$ and quantized layer $l$, respectively. $\Theta_l$ is abbreviated for all learnable parameters within the quantization block.
\subsection{Integration with the Transformer Architecture}
We integrate BATQuant into both LLM (Qwen3) and MLLM (Qwen3-VL) architectures by inserting block-wise affine transformations into the transformer block, where the weight-side transformations are merged into the linear layers offline, while the activation-side transformations are applied online during inference. Following the conventional practices, we employ low-bit matrix multiplications for all linear layers, while keeping
layer normalization layers, pre-quantization transformations,
RoPE embeddings and attention scores in BF16.
\subsubsection{MLP Module.} In LLM and the text model of MLLM, the MLP module employs two transformation sets, $\mathbf{P}_{up}$ and $\mathbf{P}_{down}$. $\mathbf{P}_{up}$ flattens the activation distribution after LayerNorm before the \texttt{up\_proj} and \texttt{gate\_proj} layers. $\mathbf{P}_{down}$ smooths the input to the \texttt{down\_proj} layer. In the ViT model of MLLM, the MLP module also employs two transformation sets: $\mathbf{P}_{fc1}$ and $\mathbf{P}_{fc2}$. $\mathbf{P}_{fc1}$ flattens the activation distribution after LayerNorm before the \texttt{linear\_fc1} layers. $\mathbf{P}_{fc2}$ smooths the input to the \texttt{linear\_fc2} layer. All matrices utilize the GPK decomposition to minimize storage. 
\subsubsection{Self-Attention Module.} In LLM and the text model of MLLM, the Self-Attention module employs four transformations: $\mathbf{P}_{qkv}$, $\mathbf{P}_{o}$, $\mathbf{P}_{k}$ and $\mathbf{P}_{v}$. $\mathbf{P}_{qkv}$ and $\mathbf{P}_{o}$ flatten the activation distribution before the \texttt{qkv\_proj} layer and \texttt{o\_proj} layer respectively. $\mathbf{P}_{k}$ and $\mathbf{P}_{v}$ are used to transform the key and value cache head by head, respectively. In the ViT model of MLLM, only $\mathbf{P}_{qkv}$ and $\mathbf{P}_{o}$ are employed. This is because ViT does not require an autoregressive KV cache mechanism. Consequently, there is no need to store, transform and quantize the key and value states across generation steps.

\label{sec:integration}
\section{Experiments}
\label{sec:exp}
\subsection{Settings}
\subsubsection{Evaluation and Baselines.} 
We evaluate BATQuant on Qwen3-VL-8B-Instruct (MLLM)~\cite{bai2025qwen3} and Qwen3-8B (LLM)~\cite{yang2025qwen3}. We assess quantized models on the following benchmarks: (1) Multimodal benchmarks, including MME~\cite{fu2025mme}, OCRBench~\cite{liu2024ocrbench}, DocVQA~\cite{mathew2021docvqa}, RealWorldQA~\cite{xAI2024RealWorldQA}, and VLMBlind. (2) Non-reasoning tasks, including PIQA~\cite{bisk2020piqa}, Winogrande~\cite{sakaguchi2021winogrande}, Hellaswag~\cite{zellers2019hellaswag}, ARC-Easy~\cite{clark2018think}, and ARC-Challenge~\cite{clark2018think}. (3) Reasoning benchmarks, including GSM8K~\cite{cobbe2021training}, MATH-500~\cite{lightman2023let}, AIME24, AIME25 and GPQA-D~\cite{rein2024gpqa}. We compare BATQuant against popular post-training quantization methods, including QuaRot~\cite{ashkboos2024quarot}, SpinQuant~\cite{liu2024spinquant}, BRQ~\cite{shao2025block}, FlatQuant~\cite{sun2024flatquant}, SmoothQuant~\cite{xiao2023smoothquant} and GPTQ~\cite{frantar2022gptq}. More details about benchmarks and baseline methods are provided in Appendix \ref{appendix:imple}.
 
\subsubsection{Implementation Details.}
We implement BATQuant based on Huggingface~\cite{wolf2020transformers}, PyTorch~\cite{paszke2019pytorch}. We adopt the AdamW optimizer with an
initial learning rate of 2e-3 and employ a cosine annealing
learning rate decay schedule. BATQuant is trained for 5 epochs, and the batch size is set to 4. For GPK, we set the size of the global shared matrix $g_1$ and block-specific private matrix $g_2$ to 8 and 4, respectively. For LLM, we use the BF16 model to self-generate data on the Numina-Math-1.5~\cite{numina_math_datasets} dataset, and randomly sample 128 text sequences of length 2048 to construct the calibration set. For MLLM, we randomly sample 128 image-text pairs from the GQA~\cite{hudson2019gqa} dataset to construct the calibration set. Further details about implementation are provided in Appendix \ref{appendix:imple}.
\subsubsection{Quantization Settings.}
We evaluate the proposed method on several MXFP-based quantization configurations, including weight-activation quantization and KV cache quantization. For clarity, we denote each configuration using the format $\text{W}\{\text{bits}\}\text{A}\{\text{bits}\}\text{KV}\{\text{bits}\}$. For example, W4A8KV8 indicates quantizing weights to 4-bit, activations to 8-bit, and KV cache to 8-bit. We empirically observe that combining different methods with GPTQ universally enhances performance. Consequently, unless otherwise specified, the reported results refer to the GPTQ-integrated variants of each method. Detailed comparisons between GPTQ and RTN weight quantizer are provided in Appendix \ref{appendix:add_result}.
\subsection{Main Results}
Here, we present a comprehensive empirical evaluation of BATQuant. Our experiments are designed to answer the following critical questions: (1) Can BATQuant maintain satisfactory performance under aggressive MXFP-based quantization configurations where existing methods fail? (2) How does our approach generalize across modalities (MLLMs vs. LLMs) and task domains, specifically spanning \textit{multimodal understanding} (including document understanding, STEM puzzles, and general VQA) in MLLMs and \textit{linguistic task} (covering non-reasoning and reasoning tasks) in LLMs? 
\subsubsection{Results on Multimodal Benchmarks.}
\begin{table*}[!t]
\caption{Performance comparison of various quantization methods on multimodal benchmarks across different bit-width configurations (e.g., W4A8KV16, W4A4KV16, W4A8KV8 and W4A8KV4).The recovery rate relative to the BF16 baseline is also provided and the best result in each case is marked in bold.
}
\renewcommand{\arraystretch}{1.0}
\centering
\resizebox{1.0\linewidth}{!}{
\begin{tabular}{llccccc|c}
\toprule[1.5pt]
\textbf{Bits} & \textbf{Method} &
\textbf{MME} &
\textbf{OCRBench} &
\textbf{DocVQA} &
\textbf{RealWorldQA} &
\textbf{VLMBlind} &
\textit{\textbf{Recovery(\%)}}  \\
\cmidrule(lr){1-7} \cmidrule(l){8-8}
BF16 & --               & 2377 &	906 &	95.81 &	70.98 &	73.98 &	100.00
 \\ 
\cmidrule{1-8} 

\multirow{8}{*}{W4A8KV16} & RTN           & 2294  & 883 & 94.72 & 69.80 & 70.99 & 97.43  \\
     & QuaRot        & 2327 & 870 & 95.07 & 69.80 & 71.12 & 97.53 \\
     % & QuaRot*       & -- &	-- &	-- &	-- &	-- &	-- &	-- & -- & -- \\
     & SpinQuant     & 2321 & 872 & 94.79 & \textbf{70.46} & 69.82 & 97.29 \\
     & BRQ     & 2329 & 865 & 94.72 & 70.19 & 67.18 & 96.40 \\
     % & SpinQuant*    & -- & -- & -- & -- & -- & -- & -- & -- &  -- \\
     & FlatQuant     & 2351 & 886 & 95.31 & 69.02 & \textbf{73.90} & 98.66 \\
     % & FlatQuant*    & -- & -- & -- & -- & -- & -- & -- &  -- \\
%      & AWQ           & -- &	-- &	-- &	-- &	-- &	-- & --
% & -- \\
     & SmoothQuant   & 2349 & 885 & 94.81 & 70.06 & 69.46 & 97.61 \\
     % & SmoothQuant*  & -- &	-- &	-- &	-- &	-- &	-- &	-- & -- & -- \\
     & GPTQ          & 2346 & 891 & 95.03 & 69.15 & 72.62 & 98.36 \\ 
     \cmidrule{2-8}
     & \cellcolor{gray!15}\textbf{BATQuant}          & \cellcolor{gray!15}\textbf{2386} & \cellcolor{gray!15}\textbf{893} & \cellcolor{gray!15}\textbf{95.55} & \cellcolor{gray!15}70.20 & \cellcolor{gray!15}{73.14} & \cellcolor{gray!15}\textbf{99.29} \\ 
\cmidrule{1-8} 

\multirow{8}{*}{W4A4KV16} & RTN           & 2243 & 838 & 92.70 & 65.23 & 66.47 & 93.07 \\
     & QuaRot        & 2189 & 810 & 93.47 & 64.97 & 57.62 & 89.69 \\
     % & QuaRot*       & -- &	-- &	-- &	-- &	-- & -- \\
     & SpinQuant     & 1994 & 801 & 91.79 & 65.36 & 60.23 & 88.32 \\
     & BRQ     & 2147 & 805 & 92.94 & 66.14 & 62.14 & 90.74 \\
     % & SpinQuant*    & -- &	-- &	-- &	-- &	-- & -- \\
     & FlatQuant     & 2231 & \textbf{873} & 94.10 & 65.62 & 68.86 & 94.79 \\
     % & FlatQuant*    & -- &	-- &	-- &	-- &	-- & -- \\
     % & AWQ           & -- &	-- &	-- &	-- &	-- & -- \\
     & SmoothQuant   & 2264 & 862 & 93.93 & \textbf{68.89} & 66.26 & 95.01 \\
     % & SmoothQuant*  & -- &	-- &	-- &	-- &	-- & -- \\
     & GPTQ          & 2286 & 849 & 93.98 & 66.93 & 67.29 & 94.64 \\ 
     \cmidrule{2-8}
     & \cellcolor{gray!15}\textbf{BATQuant}          & \cellcolor{gray!15}\textbf{2360} & \cellcolor{gray!15}864 & \cellcolor{gray!15}\textbf{94.31} & \cellcolor{gray!15}{67.32} & \cellcolor{gray!15}\textbf{69.70} & \cellcolor{gray!15}\textbf{96.43} \\ 
\cmidrule{1-8} 

\multirow{8}{*}{W4A8KV8} & RTN           & 2208 & 878 &	94.64 & 69.54 & 71.01 & 96.51 \\
     & QuaRot        & 2296 & 868 & 95.11 & 69.02 & 70.26 & 96.77 \\
     % & QuaRot*       & -- &	-- &	-- &	-- &	-- & -- \\
     & SpinQuant     & 2217 & 832 & 94.41 & 68.10 & 69.04 & 94.58 \\
     & BRQ     & 2283 & 867 & 94.63 & 69.80 & 67.36 & 95.98 \\
     % & SpinQuant*    & -- &	-- &	-- &	-- &	-- & -- \\
     & FlatQuant     & 2353 & 888 & 95.12 & 69.14 & 72.77 & 98.41 \\
     % & FlatQuant*    & -- &	-- &	-- &	-- &	-- & -- \\
     % & AWQ           & -- &	-- &	-- &	-- &	-- & -- \\
     & SmoothQuant   & 2317 & 884 & 94.72 & 70.19 & 68.91 & 97.19 \\
     % & SmoothQuant*  & -- &	-- &	-- &	-- &	-- & -- \\
     & GPTQ          & 2340 & 885 & 95.14 & \textbf{71.11} & 71.79 & 98.53 \\ 
     \cmidrule{2-8}
     & \cellcolor{gray!15}\textbf{BATQuant}          & \cellcolor{gray!15}\textbf{2368} & \cellcolor{gray!15}\textbf{890} & \cellcolor{gray!15}\textbf{95.47} & \cellcolor{gray!15}69.93 & \cellcolor{gray!15}\textbf{72.82} & \cellcolor{gray!15}\textbf{98.89} \\ 
\cmidrule{1-8} 

\multirow{8}{*}{W4A8KV4} & RTN           & 2220 & 856 & 94.05 & 68.50 & 67.50 & 94.76 \\
     & QuaRot        & 2280 & 857 & 94.66 & 68.52 & 68.36 & 95.65 \\
     % & QuaRot*       & -- &	-- &	-- &	-- &	-- & -- \\
     & SpinQuant     & 2248 &	829 &	94.18 &	68.63 &	64.50 & 93.65 \\
     & BRQ     & 2236 & 841 & 94.07 & 68.63 & 66.03 & 94.20 \\
     % & SpinQuant*    & -- &	-- &	-- &	-- & -- & -- \\
     & FlatQuant     & 2293 & 884 & 94.88 & \textbf{68.76} & 70.75 & 97.11 \\
     % & FlatQuant*    & -- &	-- &	-- &	-- 
     % & -- & --  \\
     % & AWQ           & -- &	-- &	-- &	-- &	-- & -- \\
     & SmoothQuant   & 2283 & 871 & 94.39 & 67.02 & 66.99 & 95.13  \\
     % & SmoothQuant*  & -- &	-- &	-- &	-- &	-- & --  \\
     & GPTQ          & 2328 & 867 & 94.15 & 68.10 & 70.81 & 96.71 \\ 
     \cmidrule{2-8}
     & \cellcolor{gray!15}\textbf{BATQuant}          & \cellcolor{gray!15}\textbf{2332} & \cellcolor{gray!15}\textbf{885} & \cellcolor{gray!15}\textbf{95.07} & \cellcolor{gray!15}68.63 & \cellcolor{gray!15}\textbf{70.92} & \cellcolor{gray!15}\textbf{97.51} \\ 
\bottomrule[1.5pt]
\end{tabular}
}
% \vspace{-0.25cm}
\label{tab:qwen3vl}
\end{table*}
Table~\ref{tab:qwen3vl} summarizes the performance of different post-training quantization methods on the Qwen3-VL-8B-Instruct model across five multimodal benchmarks. As shown in the table, BATQuant consistently establishes state-of-the-art results across all bit-width configurations. Notably, in the aggressive \texttt{W4A4KV16} regime, BATQuant achieves an average recovery rate of {96.43\%}, significantly outperforming the strongest baseline FlatQuant by a margin of 1.64\%. Under \texttt{W4A8KV16} scenario, BATQuant achieves an average recovery rate of {99.29\%}, which is the only approach exhibiting a performance degradation of under 1\%. This superiority extends to KV cache quantization as well. Under \texttt{W4A8KV8} and \texttt{W4A8KV4}, our method maintains superior performance with recovery rates of 98.89\% and 97.51\%, respectively. Such a consistent performance gain is also widely observed across different types of benchmarks, including document understanding, STEM puzzles, and general VQA. We attribute this success to our method's unique capability to mitigate inter-block energy transfer, thereby effectively capturing diverse outlier patterns that conventional methods fail to address.
\begin{figure}[t]
    \centering
    
    \begin{subfigure}[b]{0.49\linewidth}
        \centering
        \includegraphics[width=\linewidth]{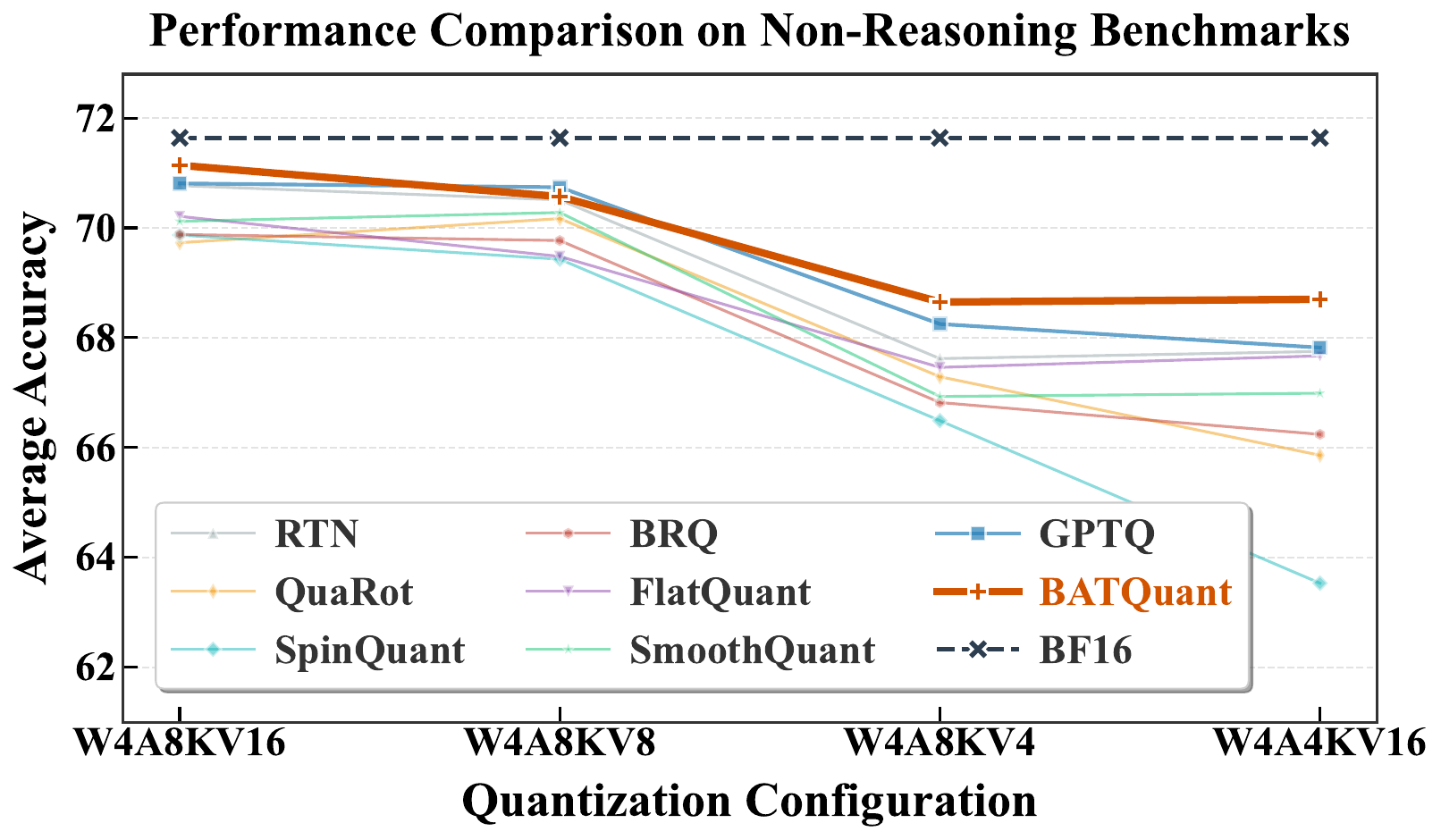}
        \caption{Performance of Qwen3-8B on Non-Reasoning tasks under different quantization settings.}
        \label{fig:non_reasoning}
    \end{subfigure}
    \hfill
    \begin{subfigure}[b]{0.49\linewidth}
        \centering
        \includegraphics[width=\linewidth]{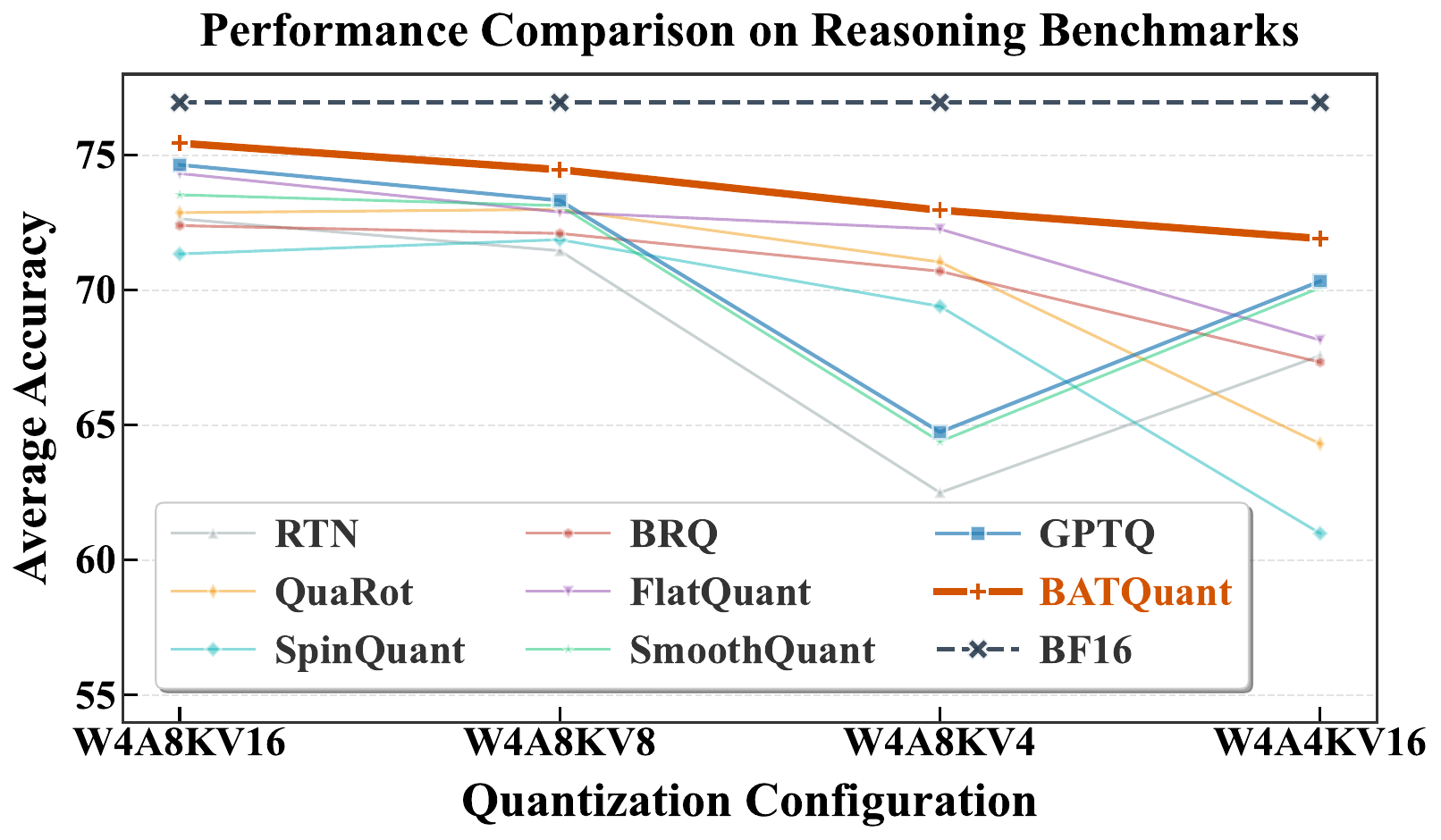}
        \caption{Performance of Qwen3-8B on Reasoning tasks under different quantization settings.}
        \label{fig:reasoning}
    \end{subfigure}
    
    \caption{Performance comparison of different methods on Qwen3-8B across LLM benchmarks under various quantization configurations. The results are categorized into Non-Reasoning (left) and Reasoning (right) tasks.}
    \label{fig:llm_benchmarks_overall}
\end{figure}
\begin{table*}[!t]
\caption{Performance comparison of various quantization methods on reasoning benchmarks across different bit-width configurations (e.g., W4A8KV16, W4A4KV16, W4A8KV8 and W4A8KV4).The recovery rate relative to the BF16 baseline is also provided and the best result in each case is marked in bold.
}
\renewcommand{\arraystretch}{1.0}
\centering
\resizebox{1.0\linewidth}{!}{
\begin{tabular}{llccccc|cc}
\toprule[1.5pt]
\textbf{Bits} & \textbf{Method} &
\textbf{GSM8K} &
\textbf{MATH-500} &
\textbf{AIME24} &
\textbf{AIME25} &
\textbf{GPQA-D} &
\textbf{Avg.} &
\textit{\textbf{Recovery(\%)}}  \\
\cmidrule(lr){1-7} \cmidrule(l){8-9}
BF16 & --               & 95.15 & 96.87 & 71.46 & 63.12 & 58.13 & 76.95 & 100.00
 \\ 
\cmidrule{1-9} 

\multirow{8}{*}{W4A8KV16} & RTN           & 93.71 & 95.53 & 64.58 & 55.00 & 54.39 & 72.64 & 93.64 \\
     & QuaRot        & 94.47 & 95.67 & 64.17 & 55.63 & 54.39 & 72.87 & 93.91 \\
     % & QuaRot*       & -- &	-- &	-- &	-- &	-- &	-- &	-- & -- & -- \\
     & SpinQuant     & 94.69 & 95.53 & 60.42 & 51.46 & 54.58 & 71.34 & 91.62 \\
     & BRQ     & 93.71 & 95.80 & 63.96 & 53.33 & 55.40 & 72.39 & 93.26 \\
     % & SpinQuant*    & -- & -- & -- & -- & -- & -- & -- & -- &  -- \\
     & FlatQuant     & 94.62 & 95.93 & \textbf{69.17} & 57.08 & 54.80 & 74.32 & 95.99 \\
     % & FlatQuant*    & -- & -- & -- & -- & -- & -- & -- &  -- \\
%      & AWQ           & -- &	-- &	-- &	-- &	-- &	-- & --
% & -- \\
     & SmoothQuant   & \textbf{94.92} & 96.27 & 65.62 & 56.04 & 54.80 & 73.53 & 94.80 \\
     % & SmoothQuant*  & -- &	-- &	-- &	-- &	-- &	-- &	-- & -- & -- \\
     & GPTQ          & 94.39 & 96.33 & 68.02 & 59.38 & 55.10 & 74.64 & 96.54 \\ 
     \cmidrule{2-9}
     & \cellcolor{gray!15}\textbf{BATQuant}          & \cellcolor{gray!15}94.84 & \cellcolor{gray!15}\textbf{96.40} & \cellcolor{gray!15}68.33 & \cellcolor{gray!15}\textbf{59.38} & \cellcolor{gray!15}\textbf{57.22} & \cellcolor{gray!15}\textbf{75.23} & \cellcolor{gray!15}\textbf{97.46} \\ 
\cmidrule{1-9} 

\multirow{8}{*}{W4A4KV16} & RTN           & 93.10 & 94.53 & 53.33 & 47.08 & 49.80 & 67.57 & 86.06 \\
     & QuaRot        & 94.09 & 92.47 & 47.50 & 39.37 & 48.13 & 64.31 & 81.20 \\
     % & QuaRot*       & -- &	-- &	-- &	-- &	-- & -- \\
     & SpinQuant     & 93.40 & 91.67 & 38.57 & 35.63 & 45.66 & 60.99 & 76.35 \\
     & BRQ     & 92.27 & 91.73 & 37.29 & 34.58 & 48.03 & 60.78 & 76.25 \\
     % & SpinQuant*    & -- &	-- &	-- &	-- &	-- & -- \\
     & FlatQuant     & 93.40 & 94.33 & 58.96 & 43.54 & 50.51 & 68.15 & 86.78 \\
     % & FlatQuant*    & -- &	-- &	-- &	-- &	-- & -- \\
     % & AWQ           & -- &	-- &	-- &	-- &	-- & -- \\
     & SmoothQuant   & 94.69 &	95.33 &	60.71 &	47.29 &	52.42 & 70.09 & 89.60 \\
     % & SmoothQuant*  & -- &	-- &	-- &	-- &	-- & -- \\
     & GPTQ          & 94.24 & \textbf{95.73} & 57.50 & 52.08 & 52.12 & 70.33 & 90.10 \\ 
     \cmidrule{2-9}
     & \cellcolor{gray!15}\textbf{BATQuant}          & \cellcolor{gray!15}\textbf{94.77} & \cellcolor{gray!15}95.60 & \cellcolor{gray!15}\textbf{62.08} & \cellcolor{gray!15}\textbf{52.92} & \cellcolor{gray!15}\textbf{54.19} & \cellcolor{gray!15}\textbf{71.91} & \cellcolor{gray!15}\textbf{92.45} \\ 
\cmidrule{1-9} 

\multirow{8}{*}{W4A8KV8} & RTN           & 93.78 & 95.00 & 60.21 & 54.79 & 53.54 & 71.46 & 91.96 \\
     & QuaRot        & 94.09 & 95.73 & 64.79 & 55.83 & 54.49 & 72.99 & 94.11 \\
     % & QuaRot*       & -- &	-- &	-- &	-- &	-- & -- \\
     & SpinQuant     & 94.47 & 95.47 & 59.38 & 53.96 & 55.86 & 71.87 & 92.56 \\
     & BRQ     & 94.69 & 95.33 & 63.75 & 52.71 & 54.04 & 72.10 & 92.72 \\
     % & SpinQuant*    & -- &	-- &	-- &	-- &	-- & -- \\
     & FlatQuant     & 94.54 & 96.00 & 65.42 & 53.96 & 54.55 & 72.89 & 93.87 \\
     % & FlatQuant*    & -- &	-- &	-- &	-- &	-- & -- \\
     % & AWQ           & -- &	-- &	-- &	-- &	-- & -- \\
     & SmoothQuant   & 94.39 & 96.13 & 66.04 & 54.79 & 54.29 & 73.13 & 94.21 \\
     % & SmoothQuant*  & -- &	-- &	-- &	-- &	-- & -- \\
     & GPTQ          & 94.47 & 96.13 & 65.00 & \textbf{57.08} & 53.94 & 73.32 & 94.54 \\ 
     \cmidrule{2-9}
     & \cellcolor{gray!15}\textbf{BATQuant}          & \cellcolor{gray!15}\textbf{94.62} & \cellcolor{gray!15}\textbf{96.27} & \cellcolor{gray!15}\textbf{69.37} & \cellcolor{gray!15}55.21 & \cellcolor{gray!15}\textbf{56.82} & \cellcolor{gray!15}\textbf{74.46} & \cellcolor{gray!15}\textbf{96.22} \\ 
\cmidrule{1-9} 

\multirow{8}{*}{W4A8KV4} & RTN           & 92.12 & 91.13 & 43.54 & 38.75 & 46.97 & 62.50 & 78.80 \\
     & QuaRot        & 94.01 & 94.80 & 62.08 & 52.50 & 51.82 & 71.04 & 91.17 \\
     % & QuaRot*       & -- &	-- &	-- &	-- &	-- & -- \\
     & SpinQuant     & 93.25 & 94.33 & 57.71 &	49.58 & 52.12  & 69.40 & 88.87 \\
     & BRQ     & 93.56 & 95.13 & 62.08 & 49.17 & 53.54 & 70.70 & 90.68 \\
     % & SpinQuant*    & -- &	-- &	-- &	-- & -- & -- \\
     & FlatQuant     & 94.09 & \textbf{95.40} & 63.33 & 53.54 & \textbf{54.95} & 72.26 & 93.07\\
     % & FlatQuant*    & -- &	-- &	-- &	-- 
     % & -- & --  \\
     % & AWQ           & -- &	-- &	-- &	-- &	-- & -- \\
     & SmoothQuant   & 93.03 & 92.73 & 46.67 & 40.33 & 49.19  & 63.39 & 81.46 \\
     % & SmoothQuant*  & -- &	-- &	-- &	-- &	-- & --  \\
     & GPTQ          & 93.40 & 93.07 & 47.92 & 39.58 & 49.75 & 64.74 & 81.92 \\ 
     \cmidrule{2-9}
     & \cellcolor{gray!15}\textbf{BATQuant}          & \cellcolor{gray!15}\textbf{94.77} & \cellcolor{gray!15}95.27 & \cellcolor{gray!15}\textbf{66.04} & \cellcolor{gray!15}\textbf{54.48} & \cellcolor{gray!15}54.24 & \cellcolor{gray!15}\textbf{72.96} & \cellcolor{gray!15}\textbf{94.00} \\ 
\bottomrule[1.5pt]
\end{tabular}
}
% \vspace{-0.25cm}
\label{tab:qwen3_rea}
\end{table*}

\subsubsection{Results on LLM Benchmarks.}
To comprehensively evaluate the generalization capability of BATQuant beyond multimodal, we conduct extensive experiments on Qwen3-8B. The overall performance trends across all configurations are shown in Figure~\ref{fig:llm_benchmarks_overall} and the detailed results for reasoning benchmarks are summarized in Table~\ref{tab:qwen3_rea}. More detailed results can be found in Appendix \ref{appendix:add_result}.
\paragraph{Non-Reasoning Tasks.} As shown in Figure~\ref{fig:llm_benchmarks_overall}, under the \texttt{W4A8KV16} configuration, our method achieves near-lossless accuracy compared to BF16 baseline. As the quantization difficulty intensifies in the aggressive \texttt{W4A4KV16} and \texttt{W4A8KV4} regimes, rotation-based methods (e.g., SpinQuant, QuaRot) suffer from severe performance degradation while our method maintains a robust level of accuracy. This suggests that our block-wise affine transformation effectively mitigates the distortion of activation distributions caused by extreme quantization, ensuring that fundamental linguistic patterns remain intact.
\paragraph{Reasoning Tasks.}
The disparity between BATQuant and baselines becomes even more pronounced on complex reasoning benchmarks requiring multi-step logical deduction and mathematical computation. As detailed in Table~\ref{tab:qwen3_rea}, reasoning tasks are inherently more sensitive to quantization noise due to the compounding effect of errors across long reasoning chains. In the \texttt{W4A8KV16} scenario, BATQuant achieves a recovery rate of 97.46\%, surpassing GPTQ by a substantial margin of 0.92\%. Notably, under \texttt{W4A4KV16} scenario, competing methods suffer from severe performance collapse on GSM8K and MATH-500, while BATQuant maintains a stable performance. In \texttt{W4A8KV8} and \texttt{W4A8KV4} scenarios, our method outperforms the strong baseline GPTQ and FlatQuant by 1.68\% and 0.93\%, respectively.

The consistent superiority of BATQuant across both multimodal tasks and complex linguistic reasoning underscores its remarkable cross-modality generalization. Our method maintains stable performance even under aggressive low-bit configurations where baselines fail. This broad effectiveness stems from the fundamental nature of our block-wise affine transformation, which dynamically aligns activation outliers and mitigates quantization noise at a granular level, independent of specific data modalities or task semantics.
\subsubsection{Qualitative Results.}
\begin{figure*}[!tbp]
\centering

\begin{subfigure}[b]{0.24\textwidth}
  \centering
  \includegraphics[width=\linewidth]{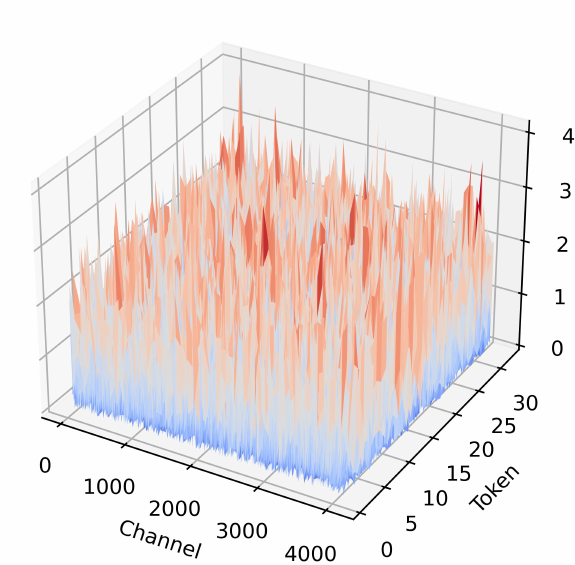}
  \caption{SpinQuant}
  \label{fig:activation_viz_spinquant}
\end{subfigure}
\hfill
\begin{subfigure}[b]{0.24\textwidth}
  \centering
  \includegraphics[width=\linewidth]{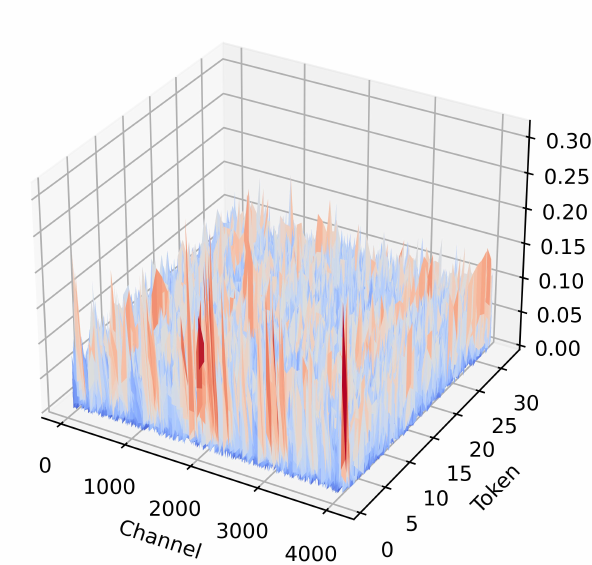}
  \caption{FlatQuant}
  \label{fig:activation_viz_flatquant}
\end{subfigure}
\hfill
\begin{subfigure}[b]{0.24\textwidth}
  \centering
  \includegraphics[width=\linewidth]{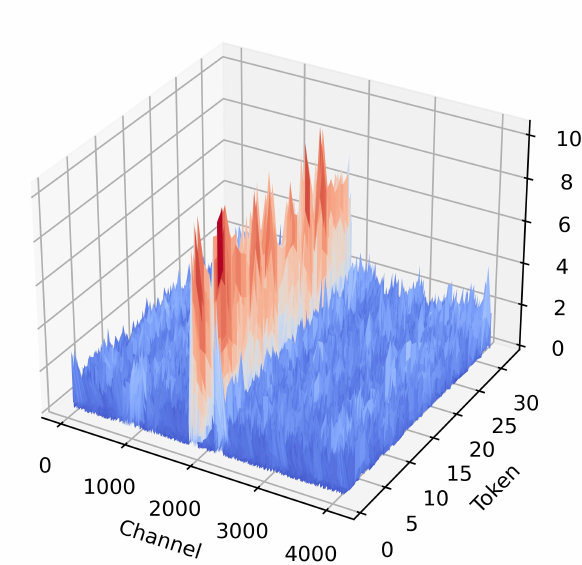}
  \caption{BRQ}
  \label{fig:activation_viz_brq}
\end{subfigure}
\hfill
\begin{subfigure}[b]{0.24\textwidth}
  \centering
  \includegraphics[width=\linewidth]{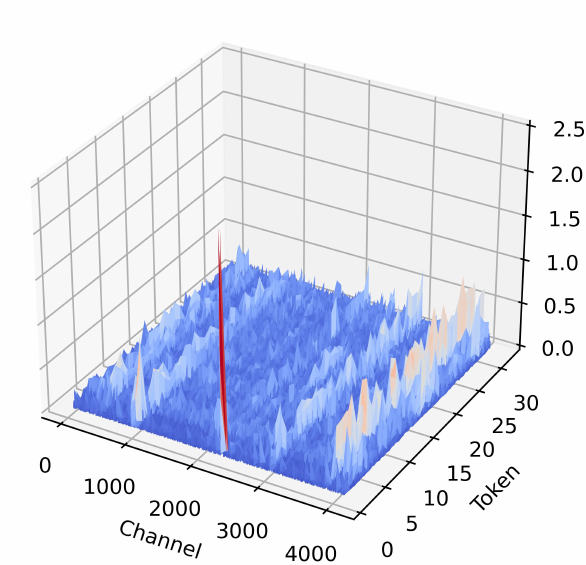}
  \caption{BATQuant (Ours)}
  \label{fig:activation_viz_ours}
\end{subfigure}
\caption{Activation distributions of the \texttt{q\_proj} module in layer 6 of Qwen3-8B with different quantization methods.}
\label{fig:activation_viz}
\end{figure*}
To provide insights into the mechanism behind our performance gains, we visualize the activation distributions across different quantization methods in Figure~\ref{fig:activation_viz}. As shown in Figure~\ref{fig:activation_viz_spinquant}, rotation-based method (e.g., SpinQuant) tend to smooth the entire tensor. While it preserves the global energy, it may transfer energy from outlier-rich blocks to smoother blocks, amplifying quantization errors in these blocks. While FlatQuant (Figure~\ref{fig:activation_viz_flatquant}) effectively suppresses global energy, it fails to prevent this inter-block energy transfer. Furthermore, although BRQ (Figure~\ref{fig:activation_viz_brq} and Figure~\ref{fig:BRQ}) introduces block-wise rotation to smooth within local blocks, our visualization reveals that it often induces a bimodal distribution within quantization blocks. Our method (Figure~\ref{fig:activation_viz_ours} and Figure~\ref{fig:BAT}) effectively prevents cross-block energy transfer while reshaping activations within blocks into a compact, unimodal distribution. More visualization results are provided in Appendix \ref{appendix:add_result}.
\subsection{Ablation Study}
To validate the effectiveness of our core designs, we conduct ablation studies on both Qwen3-8B (LLM) and Qwen3-VL-8B-Instruct (MLLM) under the \texttt{W4A4KV16} configuration. Here, we first study the effect of block-wise affine transformation and block-wise learnable clipping.
% \vspace{-10pt}

\subsubsection{Effect of Block-wise Components.}
The baseline setting without block-wise affine transformation and block-wise learnable clipping refers to the use of \textit{global-wise} counterparts. As shown in Table~\ref{tab:block_ablation}, replacing the global transformation with our block-wise variant yields significant improvements. For Qwen3-8B, applying the block-wise transformation improves the average accuracy from 68.24\% to 68.70\%. Similarly, for Qwen3-VL-8B-Instruct, it boosts the recovery rate from 95.59\% to 96.43\%. Applying block-wise clipping also provides competitive gains. For Qwen3-8B, the average accuracy is improved from 68.51\% to 68.70\%. For Qwen3-VL-8B-Instruct, the recovery rate is boosted from 96.18\% to 96.43\%. These confirm that using block-wise affine transformation and block-wise learnable clipping under MXFP quantization is crucial.
\begin{table*}[!t]
\caption{Ablation study of block-wise affine transformation and block-wise learnable clipping. We conduct the experiments under \texttt{W4A4KV16}.
% \manyi{complete the gptq variants restults}
}
\renewcommand{\arraystretch}{1.0}
\centering
\resizebox{1.0\linewidth}{!}{
\begin{tabular}{l c c c c c c c c}
\toprule
\multirow{2}{*}{\textbf{Model}} & \multicolumn{2}{c}{\textbf{Components}} & \multicolumn{5}{c}{\textbf{Non-Reasoning Benchmarks}} & \multirow{2}{*}{\textbf{Avg.}}\\
\cmidrule(lr){2-3} \cmidrule(lr){4-8}
& \rotatebox{0}{Block Trans} & \rotatebox{0}{Block Clip} & ARC-C & ARC-E & HellaSwag & PIQA & Winogrande  \\
\midrule
\multirow{3}{*}{\textbf{Qwen3-8B}} 
& \checkmark &            & 53.16 & 76.36 & 71.02 & 74.27 & \textbf{67.72} & 68.51 \\
&            & \checkmark & 52.35 & 77.44 & \textbf{71.71} & \textbf{76.01} & 63.69 & 68.24 \\
& \cellcolor{gray!15}\checkmark & \cellcolor{gray!15}\checkmark & \cellcolor{gray!15}\textbf{53.33} & \cellcolor{gray!15}\textbf{77.53} & \cellcolor{gray!15}{71.12} & \cellcolor{gray!15}{75.30} & \cellcolor{gray!15}{66.22} & \cellcolor{gray!15}\textbf{68.70} \\
\midrule
\multirow{2}{*}{\textbf{Model}} & \multicolumn{2}{c}{\textbf{Components}} & \multicolumn{5}{c}{\textbf{Multimodal Benchmarks}} & \multirow{2}{*}{\textbf{Recovery}} \\
\cmidrule(lr){2-3} \cmidrule(lr){4-8}
& Block Trans & Block Clip & MME & OCRBench & DocVQA & RealWorldQA & VLMBlind & (\%) \\
\midrule
\multirow{3}{*}{\makecell{\textbf{Qwen3-VL-8B}\\ \textbf{-Instruct}}} 
& \checkmark &            & 2235 & 861 & \textbf{94.63} & 67.19 & 69.99 & 96.18 \\
&            & \checkmark & 2249 & \textbf{865} & 94.04 & 67.21 & \textbf{70.28} & 95.59 \\
& \cellcolor{gray!15}\checkmark & \cellcolor{gray!15}\checkmark & \cellcolor{gray!15}\textbf{2360} & \cellcolor{gray!15}{864} & \cellcolor{gray!15}{94.31} & \cellcolor{gray!15}\textbf{67.32} & \cellcolor{gray!15}{69.70} & \cellcolor{gray!15}\textbf{96.43} \\
\bottomrule
\end{tabular}%
}
% \vspace{-0.25cm}
\label{tab:block_ablation}
\end{table*}
\begin{figure}[!tbp]
    \centering
    \begin{minipage}{0.49\linewidth}
        \centering
        \includegraphics[width=\linewidth]{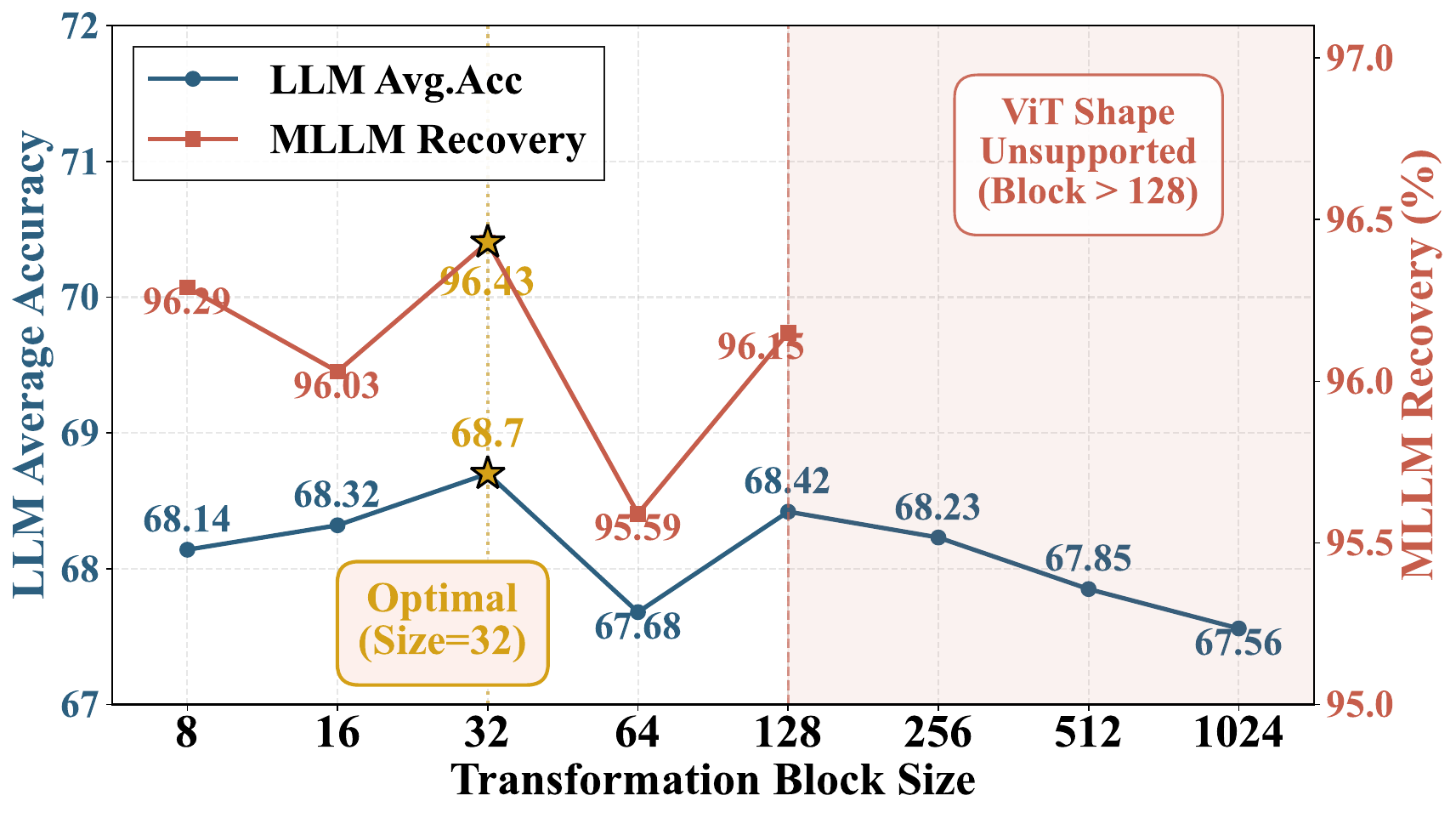}
        \caption{Performance of Qwen3-8B (LLM) and Qwen3-VL-8B-Instruct (MLLM) with different transformation block sizes.}
        \label{fig:block_size_ablation}
    \end{minipage}
    \hfill
    \begin{minipage}{0.49\linewidth}
        \centering
        % \vspace{1.5ex}
        \includegraphics[width=\linewidth]{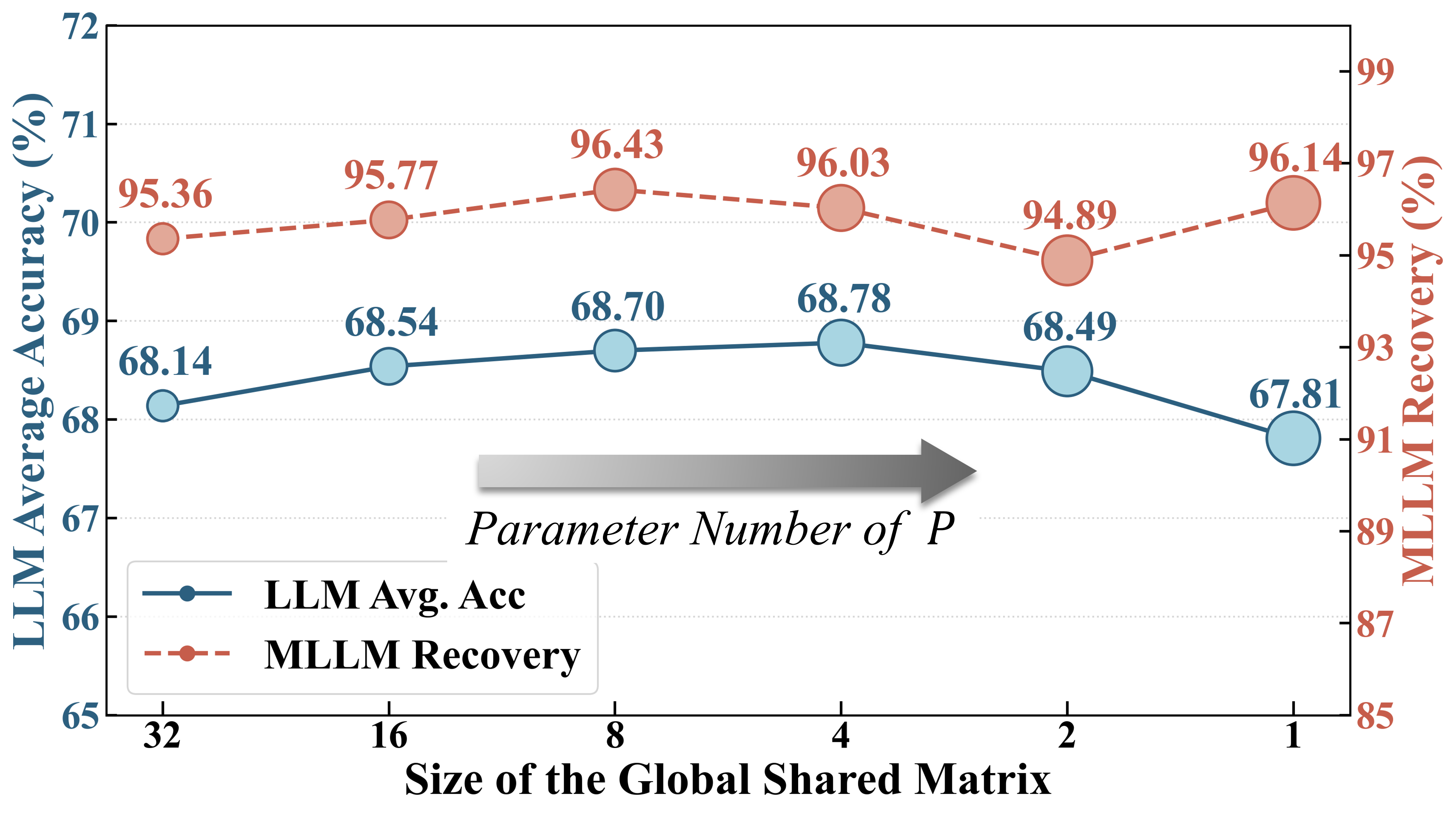}
        \caption{Performance of Qwen3-8B (LLM) and Qwen3-VL-8B-Instruct (MLLM) with different sizes of the global shared matrix.}
        \label{fig:gpk_ablation}
    \end{minipage}
\end{figure}
\subsubsection{Block Size of Affine Matrix.} BATQuant aligns the block size of affine transformation to the MXFP quantization granularity. To investigate the effect of transformation scope, we vary the size of the affine transformation $\mathbf{P}_i$ while keeping the MXFP quantization block size fixed at $g=32$. As illustrated in Figure~\ref{fig:block_size_ablation}, for Qwen3-VL-8B-Instruct and Qwen3-8B, the best performance are both achieved when the transformation block size exactly matches the quantization block size ($g=32$). This allows affine transformations to precisely reshape local distributions, isolated from cross-block outlier interference. We can also observe that deviating from this alignment leads to performance degradation. When the block size of affine matrix is smaller than $g$ (e.g., 16), the transformation scope is narrow to smooth outliers in quantization blocks. Additionally, distinct transformations lead to uneven energy ($\ell_2$-norm) suppression within quantization blocks, creating imbalanced distributions and inducing new \textit{local outliers}.
When the block size of affine matrix is greater than $g$ (e.g., 128), the transformation mixes elements across multiple quantization blocks. This transfers energy between blocks, which can increase quantization error. These findings suggest that strictly coupling the affine transformation granularity with the hardware quantization block size is an effective design choice.
\subsubsection{Effect of GPK.}To investigate the impact of Global and Private Kronecker (GPK) module, we analyze the size of the global shared matrix $\mathbf{A}$ (denoted as $g_1$). Recall that $g = g_1 \cdot g_2 = 32$; thus, varying $g_1$ inherently changes the capacity of both the shared global basis and the block-specific private components. We evaluate configurations with $g_1 \in \{1, 2, 4, 8, 16, 32\}$. The results are shown in Figure~\ref{fig:gpk_ablation}. Contrary to the intuition that increasing learnable parameters (i.e., decreasing $g_1$) monotonically improves performance, our experiments reveal a non-monotonic trend with an optimal point at $g_1=8$ or $g_1=4$. When $g_1$ is large (e.g., 16 or 32), the dimension of the private matrix $\mathbf{B}_i$ becomes small ($g_2 \le 2$), severely limiting the ability of each block to adapt its local distribution independently and leading to a performance drop. Conversely, when $g_1$ is small (e.g., 1 or 2), the number of private parameters increases significantly, theoretically offering higher capacity. However, the search space is also expanded. The optimizer may struggle to converge to a robust solution without more calibration data or hyperparameter tuning, leading to sub-optimal performance or instability. Therefore, to strike an optimal balance between accuracy and efficiency, we recommend the configuration with $g_1=8$ as the default setting.

\section{Conclusion}
In this paper, we present BATQuant, a robust framework for outlier-resilient MXFP4 quantization that leverages learnable block-wise optimization. By restricting affine transformations to align strictly with hardware quantization granularity, our method effectively eliminates the cross-block energy transfer and bimodal distributions inherent in global rotation techniques. This targeted optimization, enhanced by efficient \textit{Global and Private Kronecker} (GPK) decomposition and block-wise learnable clipping, ensures precise outlier suppression with minimal overhead. Extensive experiments on MLLMs and LLMs validate that BATQuant sets new state-of-the-art results, achieving \textbf{near-lossless} results under \texttt{W4A8KV16} and recovering up to \textbf{96.43\%} of full-precision performance under aggressive \texttt{W4A4KV16} settings. We hope this work offers a practical solution for deploying large models on emerging microscaling architectures.

\bibliographystyle{splncs04}
\bibliography{main}

\clearpage
\appendix
\section{Implementation Details}
\label{appendix:imple}
\subsection{Multimodal Benchmarks}
\begin{itemize}
    \item \textbf{MME}. It is a collection of benchmarks to evaluate the multimodal understanding capability of large vision language models (LVLMs).
    \item \textbf{OCRBench}. OCRBench is a comprehensive evaluation benchmark designed to assess the OCR capabilities of Large Multimodal Models, which contains 1000 question-answer pairs, including Text Recognition, SceneText-Centric VQA, Document-Oriented VQA, Key Information Extraction, and Handwritten Mathematical Expression Recognition.
    \item \textbf{DocVQA}. DocVQA is a benchmark for Visual Question Answering (VQA) on document images. The dataset consists of 50,000 questions defined on more than 12,000 document images.
    \item \textbf{RealWorldQA}. It is a benchmark designed to test spatial and physical reasoning. It features high-quality images taken from vehicles and egocentric views, challenging models to answer questions about object relations and environmental context in unconstrained, realistic settings.
    \item \textbf{VLMBlind}. It is a benchmark of seven novel low-level visual tasks for testing VLM ability to “see” simple geometric primitives (such as line, circles, squares, intersections) that are the basic building blocks for image tasks.
\end{itemize}
For all multimodal benchmarks, we use vllm~\cite{kwon2023efficient} backend for evaluation with a sampling temperature of 0.7, a top-p value of 0.8, a top-k value of 20 and a presence penalty of 2.0. The maximum sequence length of the model is limited to 32,768.
\subsection{Non-reasoning Benchmarks}
\begin{itemize}
    \item \textbf{PIQA}. It is a physical commonsense reasoning and corresponding benchmark dataset, which was designed to investigate the physical knowledge of existing models.
    \item \textbf{Winogrande}. Winogrande is a collection of 44k problems formulated as a fill-in-a-blank task with binary options, and the goal is to choose the right option for a given sentence, which requires commonsense reasoning.
    \item \textbf{Hellaswag}. It is a commonsense inference benchmark designed to challenge language models with adversarially filtered multiple-choice questions.
    \item \textbf{ARC-Easy \& ARC-Challenge}. The ARC dataset consists of 7,787 science exam questions drawn from a variety of sources. Each question has a multiple choice structure (typically 4 answer options). ARC-Easy contains 5,197 easy questions, and ARC-Challenge contains 2,590 hard questions.
\end{itemize}
\subsection{Reasoning Benchmarks}
\begin{itemize}
    \item \textbf{GSM8K}. GSM8K is a dataset of approximately 8,500 high-quality, linguistically diverse grade school math word problems created by human writers. We employ its test split, which contains 1,319 examples in total. We evaluate model performance using \textbf{Avg@1} (i.e., the accuracy of the first generated answer).
    \item \textbf{MATH-500}. A benchmark that contains a mix of easy and hard mathematical problems designed to test comprehensive reasoning abilities. We evaluate model performance using \textbf{Avg@3} which averages accuracy over 3 independently sampled reasoning traces per problem.
    \item \textbf{AIME24}. It contains 30 problems from the American Invitational Mathematics Examination (AIME) 2024. We report results using \textbf{Avg@16} which averages accuracy over 16 independently sampled reasoning traces per problem.
    \item \textbf{AIME25}. It contains 30 problems from the American Invitational Mathematics Examination (AIME) 2025. We report results using \textbf{Avg@16} which averages accuracy over 16 independently sampled reasoning traces per problem.
    \item \textbf{GPQA-D}. GPQA is a benchmark of graduate-level questions authored and validated by PhD experts. It is designed to be "Google-proof": highly skilled non-experts with unrestricted web access achieve only 34\% accuracy, while domain experts reach 65\% (74\% after error correction). We report results using \textbf{Avg@10} which averages accuracy over 10 independently sampled reasoning traces per problem.
\end{itemize}
For all reasoning benchmarks, we use vllm~\cite{kwon2023efficient} backend for evaluation with a sampling temperature of 0.6, a top-p value of 0.95 and a top-k value of 20. The maximum sequence length of the model is limited to 38,912.

\subsection{Baseline Methods}
\begin{itemize}
    \item \textbf{RTN}. It is the straightforward quantization strategy that maps original floating-point values without additional optimization or calibration.
    \item \textbf{QuaRot}. It uses randomized Hadamard transforms to rotate weights and activations into a space where outliers are suppressed, enabling outlier-free 4-bit quantization.
    \item \textbf{SpinQuant}. It employs orthogonal matrices optimized via the Cayley optimizer to rotate weights and activations into a space where outliers are suppressed.
    \item \textbf{BRQ}. It is equipped with block-wise rotation to prevent the energy transfer in weights and activations rotation.
    \item \textbf{FlatQuant}. It is designed to improve low-bit quantization by flattening the activation distributions using global affine matrices, specifically optimized for efficient deployment on hardware.
    \item \textbf{SmoothQuant}. It uses a diagnoal scales to smooth activation outliers by migrating the quantization difficulty from activations to weights.
    \item \textbf{GPTQ}. It is a layer-wise post-training quantization method that leverages approximate second-order information (Hessian) to  minimize quantization errors, achieving high accuracy for weight-only low-bit quantization.
\end{itemize}

\subsection{Hyperparameter Settings}
We implement BATQuant based on Huggingface~\cite{wolf2020transformers}, PyTorch~\cite{paszke2019pytorch}. We adopt the AdamW optimizer with an
initial learning rate of 2e-3 and employ a cosine annealing
learning rate decay schedule. BATQuant is trained for 5 epochs, and the batch size is set to 4. For GPK, we set the size of the global shared matrix $g_1$ and block-specific private matrix $g_2$ to 8 and 4, respectively. To simulate the quantization with MXFP format, we use the microxcaling library\footnote{https://github.com/microsoft/microxcaling} for all experiments.

\section{Detailed Algorithm Flow}
\label{app:algorithm}

In this section, we provide the detailed algorithmic implementation of BATQuant. We first formalize the efficient forward pass of the \textit{Global and Private Kronecker} (GPK) decomposition, followed by the complete calibration procedure for learning the block-wise affine transformations and clipping parameters.

\subsection{Efficient Inference via GPK Forward Pass}
To minimize runtime overhead during inference, the block-wise affine transformation $\mathbf{P}_i = \mathbf{B}_i \otimes \mathbf{A}$ is not materialized as a full dense matrix. Instead, we leverage the Kronecker product to perform the transformation efficiently without explicit matrix construction. Specifically, for the $i$-th block input vector of size $g = g_1 \cdot g_2$, the operation proceeds in three steps. First, the input vector is reshaped into a 3D matrix of dimensions $1 \times g_2 \times g_1$. Second, this matrix is multiplied by the global shared matrix $\mathbf{A} \in \mathbb{R}^{g_1 \times g_1}$ from the right and the block-specific private matrix $\mathbf{B}_i \in \mathbb{R}^{g_2 \times g_2}$ from the left; 
Finally, the resulting matrix is reshaped back to its original shape. Algorithm~\ref{alg:gpk_forward} details the vectorized implementation of this operation for a batch of inputs across all blocks.

\begin{algorithm}[t]
\caption{GPK Forward Pass (PyTorch Style)}
\label{alg:gpk_forward}
\begin{algorithmic}[1]
    \REQUIRE Input tensor $\mathbf{X} \in \mathbb{R}^{B \times S \times N}$, Global matrix $\mathbf{A} \in \mathbb{R}^{g_1 \times g_1}$, Private matrices $\{\mathbf{B}_i\}_{i=1}^k$, Quantization block size $g$.
    \ENSURE Transformed tensor $\tilde{\mathbf{X}} \in \mathbb{R}^{B \times S \times N}$.
    
    \STATE \textbf{Parameters:} Block count $k$, dims $g_1, g_2$ s.t. $N = k \cdot g_1 \cdot g_2$.
    
    \STATE Reshape $\mathbf{X}$ from $[B, S, N]$ to $[-1, k, g_2, g_1]$.
    
    % 使用自定义的灰色注释命令
    \item[] \textcolor{gray}{\textbf{1. Global Shared Transformation (PyTorch einsum)}}
    % 使用 \texttt 突出显示函数名，方程部分也用 \texttt 保持代码感
    \STATE $\tilde{\mathbf{X}} \leftarrow \texttt{einsum}(\mathbf{X}, \mathbf{A}, \text{equation}=\texttt{(...gij,jk->...gik)})$ 
    
    \item[] \textcolor{gray}{\textbf{2. Block-Specific Private Transformation (PyTorch einsum)}}
    \STATE Stack $\{\mathbf{B}_i\}$ into $\mathbf{B}_{stack} \in \mathbb{R}^{k \times g_2 \times g_2}$.
    \STATE $\tilde{\mathbf{X}} \leftarrow \texttt{einsum}(\mathbf{B}_{stack}, \tilde{\mathbf{X}}, \text{equation}=\texttt{(gij,bgjk->bgik)})$
    
    \STATE Reshape $\tilde{\mathbf{X}}$ back to $[B, S, N]$.
    
    \RETURN $\tilde{\mathbf{X}}$
\end{algorithmic}
\end{algorithm}

\subsection{BATQuant Calibration Procedure}
The calibration process aims to learn the optimal parameters $\Theta$ that minimize the difference between the full-precision layer output and the quantized output. Algorithm~\ref{alg:batquant_calibration} outlines the end-to-end training flow. For each layer in the network, we iterate over a small calibration dataset:
\begin{enumerate}
    \item In each iteration, we apply the GPK-based affine transformation to weights and activations (Line 3-4).
    \item We apply the block-wise learnable clipping to weights and activations (Line 5).
    \item The transformed activations and the corresponding inverse-transformed weights are quantized to the target MXFP format (Line 6).
    \item The loss is computed as the Mean Squared Error (MSE) between the full-precision output and the quantized output (Line 7-8).
    \item Parameters are updated via backpropagation using the AdamW optimizer (Line 9).
\end{enumerate}

After calibration, the weight-side transformation $\mathbf{P}^{-1}$ is fused into the original weights $\mathbf{W}$ offline, while the activation-side transformation $\mathbf{P}$ and clipping parameters are retained for online inference.

\begin{algorithm}[t]
\caption{BATQuant Algorithm Flow}
\label{alg:batquant_calibration}
\begin{algorithmic}[1]
\REQUIRE Full-precision weights $\mathbf{W} \in \mathbb{R}^{M \times N}$, Layer input $\mathbf{X} \in \mathbb{R}^{B \times S \times N}$, Global matrix $\mathbf{A} \in \mathbb{R}^{g_1 \times g_1}$, Private matrices $\{\mathbf{B}_i\}_{i=1}^k$, Quantization block size $g$, Epoch $E$.
\ENSURE Calibrated parameters $\Theta = \{ \mathbf{A}, \{\mathbf{B}_i\}, \{\alpha_i^{\text{min}}, \alpha_i^{\text{max}}\} \}$ for each layer.
    \FOR{epoch $= 1$ to $E$}
        \FOR{each batch in $\mathbf{X}$}
            \item[] \textcolor{gray}{\textbf{1. Transformation}}
            \STATE Obtain transformed activations $\tilde{\mathbf{X}}$ using $\mathbf{X}$, $\mathbf{A}$ and $\{\mathbf{B}_i\}$ based on Alg.~\ref{alg:gpk_forward}.
            \STATE Obtain transformed weights $\tilde{\mathbf{W}}$ using $\mathbf{W}$, $\mathbf{A}^{-1}$ and $\{\mathbf{B}_i^{-1}\}$ based on Alg.~\ref{alg:gpk_forward}.
            \STATE Apply block-wise clipping to weights $\tilde{\mathbf{W}}$ and $\tilde{\mathbf{X}}$.
            \item[] \textcolor{gray}{\textbf{2. Quantization}}
            \STATE $\tilde{\mathbf{X}} \leftarrow \mathcal{Q}(\tilde{\mathbf{X}})$, $\tilde{\mathbf{W}} \leftarrow \mathcal{Q}(\tilde{\mathbf{W}})$
            \item[] \textcolor{gray}{\textbf{3. Loss Computation \& Optimization}}
            \STATE $\tilde{\mathbf{Y}} \leftarrow \tilde{\mathbf{X}} \tilde{\mathbf{W}}^{\top}$, ${\mathbf{Y}} \leftarrow {\mathbf{X}} {\mathbf{W}}^{\top}$
            \STATE $\mathcal{L} \leftarrow \| \mathbf{Y} - \tilde{\mathbf{Y}} \|_2^2$
            \STATE Update $\Theta_l$ using $\nabla_{\Theta_l} \mathcal{L}$
        \ENDFOR
    \ENDFOR
    \item[] \textcolor{gray}{\textbf{4. Offline Fusion for Deployment}}
    \STATE Obtain transformed weights $\tilde{\mathbf{W}}$ using $\mathbf{W}$, $\mathbf{A}^{-1}$ and $\{\mathbf{B}_i^{-1}\}$ based on Alg.~\ref{alg:gpk_forward}.
    \STATE Apply block-wise clipping to weights $\tilde{\mathbf{W}}$.
    \STATE $\tilde{\mathbf{W}} \leftarrow \mathcal{Q}(\tilde{\mathbf{W}})$
    \STATE Store $\Theta = \{ \mathbf{A}, \{\mathbf{B}_i\}, \{\alpha_i^{\text{min}}, \alpha_i^{\text{max}}\} \}$ for online activation transformation.
\end{algorithmic}
\end{algorithm}

\section{Additional Results}
\label{appendix:add_result}
\subsection{Results of Non-Reasoning Tasks}
\begin{table*}[!t]
\caption{Performance comparison of various quantization methods on non-reasoning benchmarks across different bit-width configurations (e.g., W4A8KV16, W4A4KV16, W4A8KV8 and W4A8KV4).The recovery rate relative to the BF16 baseline is also provided and the best result in each case is marked in bold.
}
\renewcommand{\arraystretch}{1.0}
\centering
\resizebox{1.0\linewidth}{!}{
\begin{tabular}{llccccc|cc}
\toprule[1.5pt]
\textbf{Bits} & \textbf{Method} &
\textbf{ARC-C} &
\textbf{ARC-E} &
\textbf{HellaSwag} &
\textbf{PIQA} &
\textbf{Winogrande} &
\textbf{Avg.} &
\textit{\textbf{Recovery(\%)}}  \\
\cmidrule(lr){1-7} \cmidrule(l){8-9}
BF16 & --               & 56.48 & 81.06 & 74.96 & 77.69 & 68.03 & 71.64 & 100.00
 \\ 
\cmidrule{1-9} 

\multirow{8}{*}{W4A8KV16} & RTN           & 55.72 & \textbf{80.81} & \textbf{73.29} & 77.09 & 66.93 & 70.77 & 98.75 \\
     & QuaRot        & 55.20 & 78.70 & 72.77 & 76.88 & 65.11 & 69.73 & 97.31 \\
     % & QuaRot*       & -- &	-- &	-- &	-- &	-- &	-- &	-- & -- & -- \\
     & SpinQuant     & 54.69 & 76.98 & 72.76 & \textbf{78.07} & 66.85 & 69.87 & 97.52 \\
     & BRQ     & 53.67 & 78.87 & 73.27 & 76.66 & 66.93 & 69.88 & 97.43 \\
     % & SpinQuant*    & -- & -- & -- & -- & -- & -- & -- & -- &  -- \\
     & FlatQuant     & 55.72 & 79.63 & 72.66 & 76.82 & 66.22 & 70.21 & 98.01 \\
     % & FlatQuant*    & -- & -- & -- & -- & -- & -- & -- &  -- \\
%      & AWQ           & -- &	-- &	-- &	-- &	-- &	-- & --
% & -- \\
     & SmoothQuant   & 55.80 & 79.04 & 72.38 & 76.55 & 66.85 & 70.12 & 97.93 \\
     % & SmoothQuant*  & -- &	-- &	-- &	-- &	-- &	-- &	-- & -- & -- \\
     & GPTQ          & 55.89 & 80.60 & 73.16 & 77.31 & 67.09 & 70.81 & 98.82 \\ 
     \cmidrule{2-9}
     & \cellcolor{gray!15}\textbf{Ours}          & \cellcolor{gray!15}\textbf{56.14} & \cellcolor{gray!15}79.92 & \cellcolor{gray!15}73.10 & \cellcolor{gray!15}77.97 & \cellcolor{gray!15}\textbf{68.59} & \cellcolor{gray!15}\textbf{71.14} & \cellcolor{gray!15}\textbf{99.34} \\ 
\cmidrule{1-9} 

\multirow{8}{*}{W4A4KV16} & RTN           & 52.47 & 76.89 & 70.44 & 74.16 & 64.80 & 67.75 & 94.49 \\
     & QuaRot        & 50.43 & 74.28 & 67.55 & 73.67 & 63.38 & 65.86 & 91.81 \\
     % & QuaRot*       & -- &	-- &	-- &	-- &	-- & -- \\
     & SpinQuant     & 45.65 & 68.18 & 67.41 & 74.21 & 62.19 & 63.53 & 88.36 \\
     & BRQ     & 48.55 & 74.71 & 68.79 & 75.24 & 63.93 & 66.24 & 92.14 \\
     % & SpinQuant*    & -- &	-- &	-- &	-- &	-- & -- \\
     & FlatQuant     & 50.60 & \textbf{78.20} & 70.36 & 75.63 & 63.54 & 67.67 & 94.13 \\
     % & FlatQuant*    & -- &	-- &	-- &	-- &	-- & -- \\
     % & AWQ           & -- &	-- &	-- &	-- &	-- & -- \\
     & SmoothQuant   & 50.09 & 75.72 & 70.15 & 74.37 & 64.64 & 66.99 & 93.29 \\
     % & SmoothQuant*  & -- &	-- &	-- &	-- &	-- & -- \\
     & GPTQ          & 51.28 & 76.98 & 70.47 & \textbf{75.79} & 64.56 & 67.82 & 94.44 \\ 
     \cmidrule{2-9}
     & \cellcolor{gray!15}\textbf{Ours}          & \cellcolor{gray!15}\textbf{53.33} & \cellcolor{gray!15}77.53 & \cellcolor{gray!15}\textbf{71.12} & \cellcolor{gray!15}75.30 & \cellcolor{gray!15}\textbf{66.22} & \cellcolor{gray!15}\textbf{68.70} & \cellcolor{gray!15}\textbf{95.84} \\ 
\cmidrule{1-9} 

\multirow{8}{*}{W4A8KV8} & RTN           & 55.72 & 80.51 & 72.86 & 76.55 & 66.93 & 70.51 & 98.42 \\
     & QuaRot        & 55.38 & 79.84 & 72.54 & 76.88 & 66.22 & 70.17 & 97.92 \\
     % & QuaRot*       & -- &	-- &	-- &	-- &	-- & -- \\
     & SpinQuant     & 53.50 & 77.65 & 72.56 & \textbf{77.53} & 65.9 & 69.43 & 96.80 \\
     & BRQ     & 52.99 & 78.11 & 73.09 & 76.88 & 67.8 & 69.77 & 97.26 \\
     % & SpinQuant*    & -- &	-- &	-- &	-- &	-- & -- \\
     & FlatQuant     & 52.56 & 77.10 & 72.46 & 77.09 & \textbf{68.19} & 69.48 & 96.86 \\
     % & FlatQuant*    & -- &	-- &	-- &	-- &	-- & -- \\
     % & AWQ           & -- &	-- &	-- &	-- &	-- & -- \\
     & SmoothQuant   & 55.03 & 79.21 & 72.76 & 76.99 & 67.40 & 70.28 & 98.08 \\
     % & SmoothQuant*  & -- &	-- &	-- &	-- &	-- & -- \\
     & GPTQ          & \textbf{56.06} & \textbf{80.68} & 72.95 & 77.53 & 66.46 & \textbf{70.74} & \textbf{98.72} \\ 
     \cmidrule{2-9}
     & \cellcolor{gray!15}\textbf{Ours}          & \cellcolor{gray!15}55.63 & \cellcolor{gray!15}79.80 & \cellcolor{gray!15}\textbf{73.15} & \cellcolor{gray!15}77.09 & \cellcolor{gray!15}67.17 & \cellcolor{gray!15}70.57 & \cellcolor{gray!15}98.50 \\ 
\cmidrule{1-9} 

\multirow{8}{*}{W4A8KV4} & RTN           & 51.96 & 76.89 & 70.54 & 75.08 & 63.61 & 67.62 & 94.22 \\
     & QuaRot        & 52.73 & 76.47 & 70.15 & 74.81 & 62.27 & 67.29 & 93.82 \\
     % & QuaRot*       & -- &	-- &	-- &	-- &	-- & -- \\
     & SpinQuant     & 49.32 & 74.07 & 69.82 & 75.95 & 63.30 & 66.49 & 92.53 \\
     & BRQ     & 50.68 & 75.97 & 70.38 & 74.65 & 62.43 & 66.82 & 93.04 \\
     % & SpinQuant*    & -- &	-- &	-- &	-- & -- & -- \\
     & FlatQuant     & 52.13 & 77.90 & 69.60 & 75.14 & 62.51 & 67.46 & 93.97 \\
     % & FlatQuant*    & -- &	-- &	-- &	-- 
     % & -- & --  \\
     % & AWQ           & -- &	-- &	-- &	-- &	-- & -- \\
     & SmoothQuant   & 49.74 & 73.23 & 69.61 & 75.24 & \textbf{66.85} & 66.93 & 93.28 \\
     % & SmoothQuant*  & -- &	-- &	-- &	-- &	-- & --  \\
     & GPTQ          & 52.39 & 76.52 & \textbf{71.25} & 75.73 & 65.35 & 68.25 & 95.15 \\ 
     \cmidrule{2-9}
     & \cellcolor{gray!15}\textbf{Ours}          & \cellcolor{gray!15}\textbf{53.33} & \cellcolor{gray!15}\textbf{78.54} & \cellcolor{gray!15}69.53 & \cellcolor{gray!15}\textbf{76.66} & \cellcolor{gray!15}65.19 & \cellcolor{gray!15}\textbf{68.65} & \cellcolor{gray!15}\textbf{95.71} \\ 
\bottomrule[1.5pt]
\end{tabular}
}
% \vspace{-0.25cm}
\label{tab:qwen3_nonrea}
\end{table*}
Table~\ref{tab:qwen3_nonrea} presents the comprehensive performance comparison on non-reasoning benchmarks (ARC-C, ARC-E, HellaSwag, PIQA, and Winogrande) under four distinct quantization configurations. In the most challenging \texttt{W4A4KV16} configuration, BATQuant achieves an average accuracy of \textbf{68.70\%}, corresponding to a \textbf{95.84\%} recovery rate relative to the BF16 baseline. This significantly outperforms the strongest competing methods, including GPTQ (67.82\%, 94.44\%) and FlatQuant (67.67\%, 94.13\%). Notably, rotation-based methods like SpinQuant suffer from catastrophic failure in this regime, dropping to only 63.53\% accuracy. Similarly, under the \texttt{W4A8KV4} setting with aggressive KV cache quantization, BATQuant secures the highest average accuracy (\textbf{68.65\%}) and recovery rate (\textbf{95.71\%}), surpassing GPTQ by a margin of 0.40\%. Under the \texttt{W4A8KV16} configuration, BATQuant achieves a near-lossless recovery rate of \textbf{99.34\%} (Avg. 71.14\%), establishing a new state-of-the-art result that exceeds GPTQ (98.82\%) and RTN (98.75\%). In the \texttt{W4A8KV8} setting, the performance gap narrows as the quantization difficulty decreases. Here, GPTQ achieves the highest average score (70.74\%), while BATQuant remains highly competitive with 70.57\%, outperforming all other transformation-based methods (e.g., FlatQuant at 69.48\%). 
\subsection{Results of GPTQ and RTN weight quantizer}
\begin{table}[!t]
\centering
\caption{Performance comparison of different quantization methods on multimodal benchmarks using \textbf{RTN} and \textbf{GPTQ} as weight quantizers. \textbf{Bold} indicates the best result within each quantizer setting (RTN or GPTQ) for a specific bit configuration.}
\label{tab:qwen3vl_gptqvsrtn}
\resizebox{\linewidth}{!}{%
\begin{tabular}{l l l c c c c c c}
\toprule
\multirow{2}{*}{\textbf{Bits}} & \multirow{2}{*}{\textbf{Method}} & \multirow{2}{*}{\textbf{Quantizer}} & \multicolumn{5}{c}{\textbf{Multimodal Benchmark}} & \multirow{2}{*}{\textbf{Recovery}} \\
\cmidrule(lr){4-8}
& & & \textbf{MME} & \textbf{OCRBench} & \textbf{DocVQA} & \textbf{RealWorldQA} & \textbf{VLMBlind} & (\%) \\
\midrule

% --- Block 1: W4A8KV16 ---
\multirow{8}{*}{W4A8KV16} 
& \cellcolor{rtnbg}QuaRot    & \multirow{4}{*}{RTN}  & 2201 & 814 & 93.11 & 65.36 & 63.21 & 91.43 \\
& \cellcolor{rtnbg}BRQ       &                       & 2272    & 831      & 93.66     & \textbf{69.80}     & 62.11      & 93.47 \\
& \cellcolor{rtnbg}FlatQuant &                       & 2311 & \textbf{880} & \textbf{94.65} & 66.14 & 67.96  & 95.64 \\
& \cellcolor{rtnbg}\textbf{BATQuant}  &                       & \textbf{2312} & {877} & {94.58} & {66.80} & \textbf{69.27} & \textbf{96.11} \\
\cmidrule(lr){2-9}
& \cellcolor{gptqbg}QuaRot    & \multirow{4}{*}{GPTQ} & 2327 & 870 & 95.07 & 69.80 & 71.12 & 97.53 \\
& \cellcolor{gptqbg}BRQ       &                       & 2329 & 865 & 94.72 & 70.19 & 67.18 & 96.40 \\
& \cellcolor{gptqbg}FlatQuant &                       & 2351 & 886 & 95.31 & 69.02 & \textbf{73.90} & 98.66 \\
& \cellcolor{gptqbg}\textbf{BATQuant}  &                       & \textbf{2386} & \textbf{893} & \textbf{95.55} & \textbf{70.20} & {73.14} & \textbf{99.29} \\
\midrule

% --- Block 2: W4A4KV16 ---
\multirow{8}{*}{W4A4KV16} 
& \cellcolor{rtnbg}QuaRot    & \multirow{4}{*}{RTN}  & 1965 & 710      & 90.91     & 60.48     & 55.31      & 83.18 \\
& \cellcolor{rtnbg}BRQ       &                       & 2096    & 749      & 91.09     & 61.83     & 56.65      & 85.92 \\
& \cellcolor{rtnbg}FlatQuant &                       & 2147 & \textbf{846} & 93.14 & 62.48 & 65.49      & 91.49 \\
& \cellcolor{rtnbg}\textbf{BATQuant}  &                       & \textbf{2255} & {838} & \textbf{93.68} & \textbf{64.71} & \textbf{66.84} & \textbf{93.33} \\
\cmidrule(lr){2-9}
& \cellcolor{gptqbg}QuaRot    & \multirow{4}{*}{GPTQ} & 2189 & 810 & 93.47 & 64.97 & 57.62 & 89.69 \\
& \cellcolor{gptqbg}BRQ       &                       & 2147 & 805 & 92.94 & 66.14 & 62.41 & 90.75 \\
& \cellcolor{gptqbg}FlatQuant &                       & 2231 & \textbf{873} & 94.10 & 65.62 & 68.86 & 94.79 \\
& \cellcolor{gptqbg}\textbf{BATQuant}  &                       & \textbf{2360} & {864} & \textbf{94.31} & \textbf{67.32} & \textbf{69.70} & \textbf{96.43} \\
\midrule

% --- Block 3: W4A8KV8 ---
\multirow{8}{*}{W4A8KV8} 
& \cellcolor{rtnbg}QuaRot    & \multirow{4}{*}{RTN}  & 2143 & 816      & 93.27     & 65.36     & 62.49      & 90.82 \\
& \cellcolor{rtnbg}BRQ       &                       & 2277    & 815      & 93.55     & \textbf{69.93}     & 60.24      & 92.67 \\
& \cellcolor{rtnbg}FlatQuant &                       & 2285 & \textbf{871} & 94.11 & 60.52 & 70.04  & 94.09 \\
& \cellcolor{rtnbg}BATQuant  &                       & \textbf{2301}       & 867      & \textbf{94.72}     & 66.67     & \textbf{72.71}      & \textbf{96.71} \\ % 原数据为空/0
\cmidrule(lr){2-9}
& \cellcolor{gptqbg}QuaRot    & \multirow{4}{*}{GPTQ} & 2296 & 868 & 95.11 & 69.02 & 70.26 & 96.78 \\
& \cellcolor{gptqbg}BRQ       &                       & 2283 & 867 & 94.63 & 69.80 & 67.36 & 95.98 \\
& \cellcolor{gptqbg}FlatQuant &                       & 2353 & 888 & 95.12 & 69.14 & 72.77 & 98.41 \\
& \cellcolor{gptqbg}\textbf{BATQuant}  &                       & \textbf{2368} & \textbf{890} & \textbf{95.47} & \textbf{69.93} & \textbf{72.82} & \textbf{98.89} \\
\midrule

% --- Block 4: W4A8KV4 ---
\multirow{8}{*}{W4A8KV4} 
& \cellcolor{rtnbg}QuaRot    & \multirow{4}{*}{RTN}  & 2112 & 781      & 92.67     & 62.48     & 60.34   & 88.27 \\
& \cellcolor{rtnbg}BRQ       &                       & 2194    & 807      & 92.75     & \textbf{66.27}     & 57.31      & 89.80 \\
& \cellcolor{rtnbg}FlatQuant &                       & 2257 & 867 & 94.05 & 59.87 & 71.05      & 93.84 \\
& \cellcolor{rtnbg}\textbf{BATQuant}  &                       & \textbf{2289} & \textbf{874} & \textbf{94.64} & {64.97} & \textbf{71.06} & \textbf{95.83} \\
\cmidrule(lr){2-9}
& \cellcolor{gptqbg}QuaRot    & \multirow{4}{*}{GPTQ} & 2280 & 857 & 94.66 & 68.52 & 68.36 & 95.65 \\
& \cellcolor{gptqbg}BRQ       &                       & 2236 & 841 & 94.07 & 68.63 & 66.03 & 94.21 \\
& \cellcolor{gptqbg}FlatQuant &                       & 2293 & 884 & 94.88 & \textbf{68.76} & 70.75 & 97.11 \\
& \cellcolor{gptqbg}\textbf{BATQuant}  &                       & \textbf{2332} & \textbf{885} & \textbf{95.07} & {68.63} & \textbf{70.92} & \textbf{97.51} \\
\bottomrule
\end{tabular}
}
\end{table}
\begin{table}[!t]
\centering
\caption{Performance comparison of different quantization methods on LLM non-reasoning benchmarks using \textbf{RTN} and \textbf{GPTQ} as weight quantizers. \textbf{Bold} indicates the best result within each quantizer setting (RTN or GPTQ) for a specific bit configuration.}
\label{tab:qwen3_gptqvsrtn}
\resizebox{\linewidth}{!}{% 自动缩放表格宽度以适应页面
\begin{tabular}{l l l c c c c c c}
\toprule
\multirow{2}{*}{\textbf{Bits}} & \multirow{2}{*}{\textbf{Method}} & \multirow{2}{*}{\textbf{Quantizer}} & \multicolumn{5}{c}{\textbf{Non-Reasoning Benchmark}} & \multirow{2}{*}{\textbf{Avg.}} \\
\cmidrule(lr){4-8}
& & & \textbf{ARC-C} & \textbf{ARC-E} & \textbf{HellaSwag} & \textbf{PIQA} & \textbf{Winogrande} & \\
\midrule

% --- Block 1: W4A8KV16 ---
\multirow{8}{*}{W4A8KV16} 
& \cellcolor{rtnbg}QuaRot    & \multirow{4}{*}{RTN}  & 51.37 & 75.76 & 70.04 & 76.61 & 65.67 & 67.89 \\
& \cellcolor{rtnbg}BRQ       &                       & 47.44 & 72.87 & 71.37 & 75.84 & 65.19 & 66.54 \\
& \cellcolor{rtnbg}FlatQuant &                       & \textbf{55.63} & \textbf{78.83} & \textbf{72.46} & 76.22 & 66.85 & \textbf{70.00} \\
& \cellcolor{rtnbg}\textbf{BATQuant}  &                       & 54.33 & 77.48 & 72.23 & \textbf{76.61} & \textbf{68.25} & 69.78 \\
\cmidrule(lr){2-9} % 分隔 RTN 和 GPTQ
& \cellcolor{gptqbg}QuaRot    & \multirow{4}{*}{GPTQ} & 55.20 & 78.70 & 72.77 & 76.88 & 65.11 & 69.73 \\
& \cellcolor{gptqbg}BRQ       &                       & 53.67 & 78.87 & \textbf{73.27} & 76.66 & 66.93 & 69.88 \\
& \cellcolor{gptqbg}FlatQuant &                       & 55.72 & 79.63 & 72.66 & 76.82 & 66.22 & 70.21 \\
& \cellcolor{gptqbg}\textbf{BATQuant}  &                       & \textbf{56.14} & \textbf{79.92} & {73.10} & \textbf{77.97} & \textbf{68.59} & \textbf{71.14} \\
\midrule

% --- Block 2: W4A4KV16 ---
\multirow{8}{*}{W4A4KV16} 
& \cellcolor{rtnbg}QuaRot    & \multirow{4}{*}{RTN}  & 44.88 & 70.37 & 65.09 & 74.54 & 62.51 & 63.48 \\
& \cellcolor{rtnbg}BRQ       &                       & 45.90 & 67.51 & 68.47 & 74.16 & 61.40 & 63.49 \\
& \cellcolor{rtnbg}FlatQuant &                       & \textbf{51.11} & \textbf{75.93} & 69.02 & 74.92 & 62.83 & 66.76 \\
& \cellcolor{rtnbg}\textbf{BATQuant}  &                       & {50.09} & {75.55} & \textbf{71.00} & \textbf{75.19} & \textbf{66.85} & \textbf{67.74} \\
\cmidrule(lr){2-9}
& \cellcolor{gptqbg}QuaRot    & \multirow{4}{*}{GPTQ} & 50.43 & 74.28 & 67.55 & 73.67 & 63.38 & 65.86 \\
& \cellcolor{gptqbg}BRQ       &                       & 48.55 & 74.71 & 68.79 & 75.24 & 63.93 & 66.24 \\
& \cellcolor{gptqbg}FlatQuant &                       & 50.60 & \textbf{78.20} & 70.36 & \textbf{75.63} & 63.54 & 67.67 \\
& \cellcolor{gptqbg}\textbf{BATQuant}  &                       & \textbf{53.33} & {77.53} & \textbf{71.12} & {75.30} & \textbf{66.22} & \textbf{68.70} \\
\midrule

% --- Block 3: W4A8KV4 ---
\multirow{8}{*}{W4A8KV4} 
& \cellcolor{rtnbg}QuaRot    & \multirow{4}{*}{RTN}  & 47.18 & 72.64 & 67.43 & 74.32 & 60.06 & 64.33 \\
& \cellcolor{rtnbg}BRQ       &                       & 45.82 & 69.82 & 69.71 & 74.21 & 62.43 & 64.40 \\
& \cellcolor{rtnbg}FlatQuant &                       & 48.12 & 73.23 & 68.96 & 74.37 & 63.30 & 65.60 \\
& \cellcolor{rtnbg}\textbf{BATQuant}  &                       & \textbf{50.85} & \textbf{75.97} & \textbf{70.07} & \textbf{76.50} & \textbf{64.56} & \textbf{67.59} \\
\cmidrule(lr){2-9}
& \cellcolor{gptqbg}QuaRot    & \multirow{4}{*}{GPTQ} & 52.73 & 76.47 & 70.15 & 74.81 & 62.27 & 67.29 \\
& \cellcolor{gptqbg}BRQ       &                       & 50.68 & 75.97 & 70.38 & 74.65 & 62.43 & 66.82 \\
& \cellcolor{gptqbg}FlatQuant &                       & 52.13 & 77.90 & \textbf{69.60} & 75.14 & 62.51 & 67.46 \\
& \cellcolor{gptqbg}\textbf{BATQuant}  &                       & \textbf{53.33} & \textbf{78.54} & {69.53} & \textbf{76.66} & \textbf{65.19} & \textbf{68.65} \\
\midrule

% --- Block 4: W4A8KV8 ---
\multirow{8}{*}{W4A8KV8} 
& \cellcolor{rtnbg}QuaRot    & \multirow{4}{*}{RTN}  & 52.30 & 76.47 & 69.68 & \textbf{77.04} & \textbf{65.67} & 68.23 \\
& \cellcolor{rtnbg}BRQ       &                       & 48.55 & 72.47 & 71.84 & 76.66 & 64.96 & 66.90 \\
& \cellcolor{rtnbg}FlatQuant &                       & 52.73 & 77.09 & 72.18 & 76.71 & 64.25 & 68.59 \\
& \cellcolor{rtnbg}\textbf{BATQuant}  &                       & \textbf{54.52} & \textbf{79.55} & \textbf{72.20} & {76.50} & {65.59} & \textbf{69.67} \\
\cmidrule(lr){2-9}
& \cellcolor{gptqbg}QuaRot    & \multirow{4}{*}{GPTQ} & 55.38 & \textbf{79.84} & 72.54 & 76.88 & 66.22 & 70.17 \\
& \cellcolor{gptqbg}BRQ       &                       & 52.99 & 78.11 & 73.09 & 76.88 & 67.80 & 69.77 \\
& \cellcolor{gptqbg}FlatQuant &                       & 52.56 & 77.10 & 72.46 & 77.09 & \textbf{68.19} & 69.48 \\
& \cellcolor{gptqbg}\textbf{BATQuant}  &                       & \textbf{55.63} & {79.80} & \textbf{73.15} & \textbf{77.09} & {67.17} & \textbf{70.57} \\
\bottomrule
\end{tabular}
}
\end{table}
Table~\ref{tab:qwen3vl_gptqvsrtn} and Table~\ref{tab:qwen3_gptqvsrtn} compare GPTQ and RTN as weight quantizers across various MXFP configurations. The results show that GPTQ consistently outperforms RTN in all evaluated settings. This improvement is attributed to GPTQ's approximate second-order optimization, which minimizes quantization error by accounting for inter-channel weight correlations. In contrast, RTN applies per-element rounding independently, without leveraging the structural redundancy that GPTQ utilizes for error compensation. Given these consistent results, GPTQ serves as a more effective weight quantization strategy than RTN.
\subsection{Activation Visualization}
\begin{figure}[!tbp]
    \centering
    \includegraphics[width=0.99\linewidth]{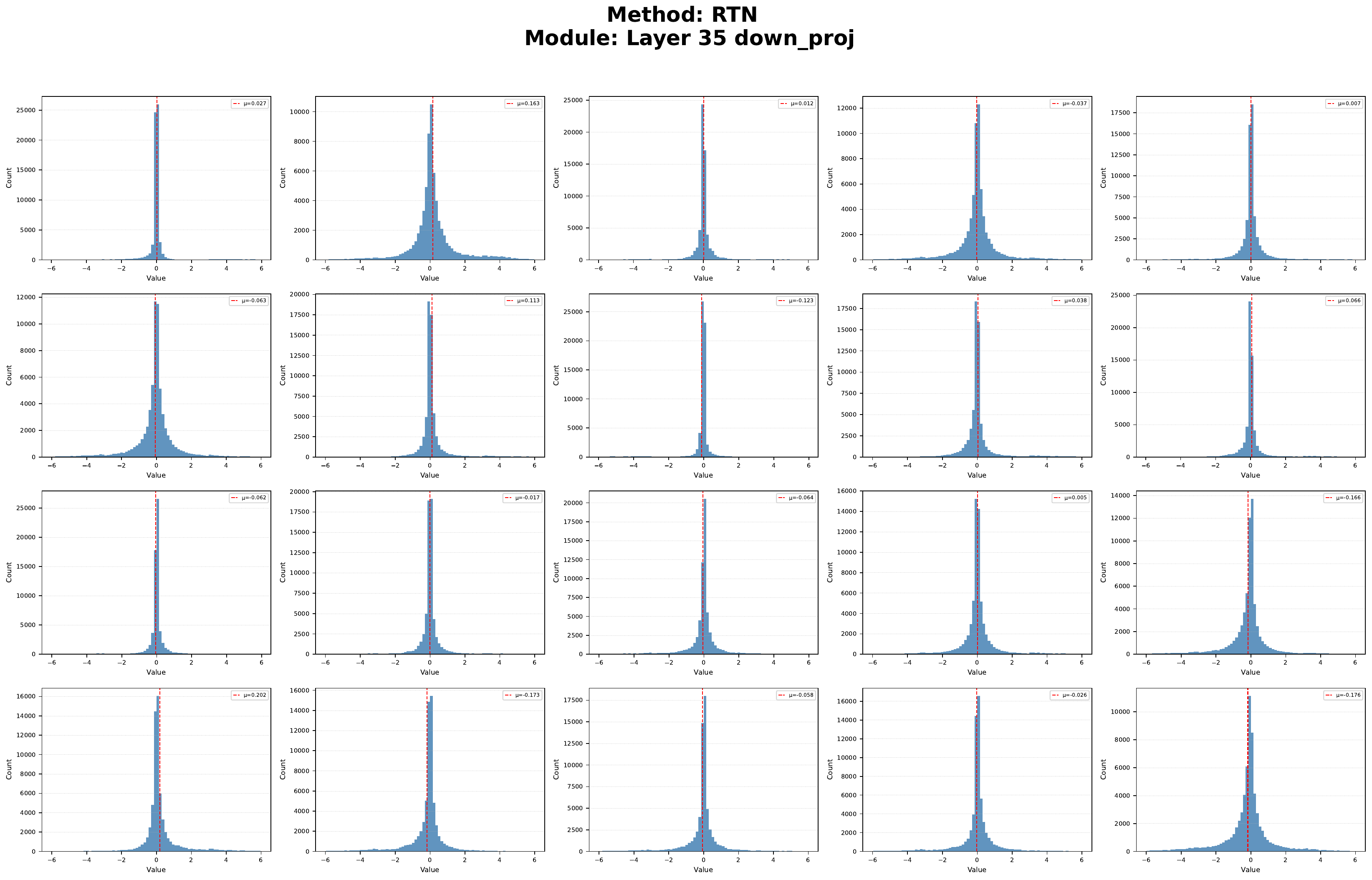}
    \caption{Activation distributions within different quantization blocks of the \texttt{down\_proj} module in layer 35 of Qwen3-8B with RTN.}
    \label{fig:activation_viz_rtn_all}
\end{figure}

\begin{figure}[!tbp]
    \centering
    \includegraphics[width=0.99\linewidth]{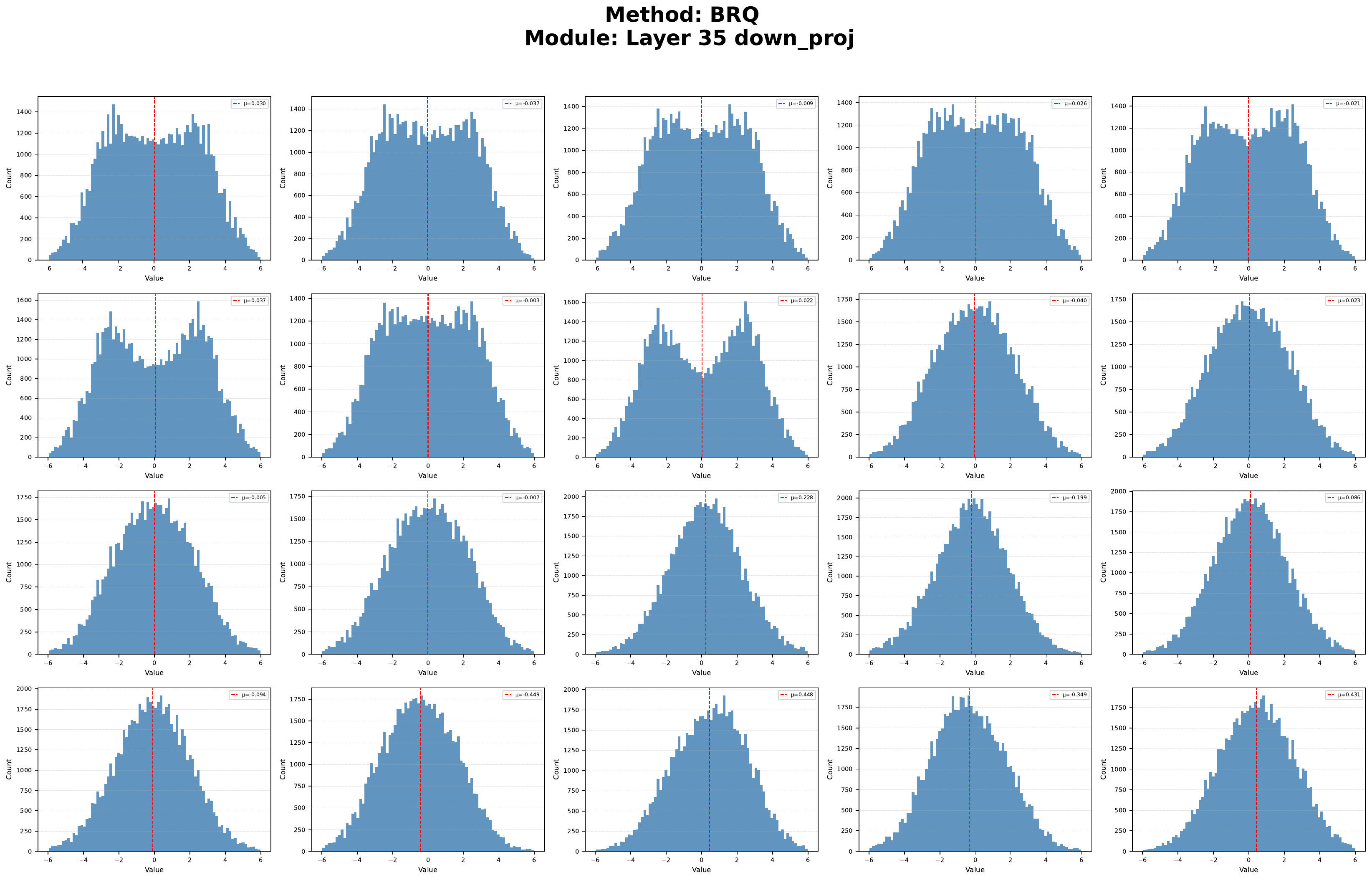}
    \caption{Activation distributions within different quantization blocks of the \texttt{down\_proj} module in layer 35 of Qwen3-8B with BRQ.}
    \label{fig:activation_viz_brq_all}
\end{figure}

\begin{figure}[!tbp]
    \centering
    \includegraphics[width=0.99\linewidth]{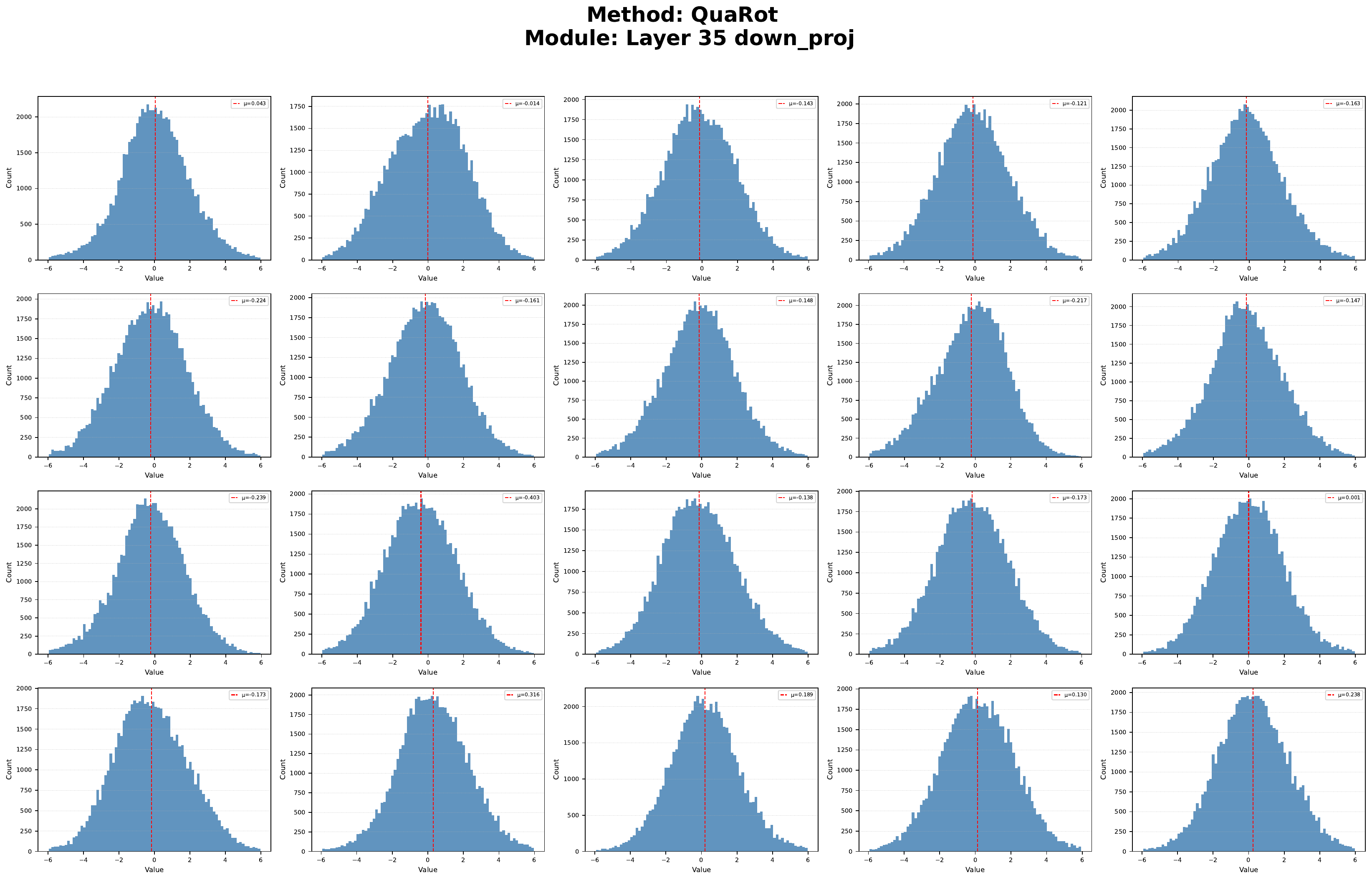}
    \caption{Activation distributions within different quantization blocks of the \texttt{down\_proj} module in layer 35 of Qwen3-8B with QuaRot.}
    \label{fig:activation_viz_quarot_all}
\end{figure}

\begin{figure}[!tbp]
    \centering
    \includegraphics[width=0.99\linewidth]{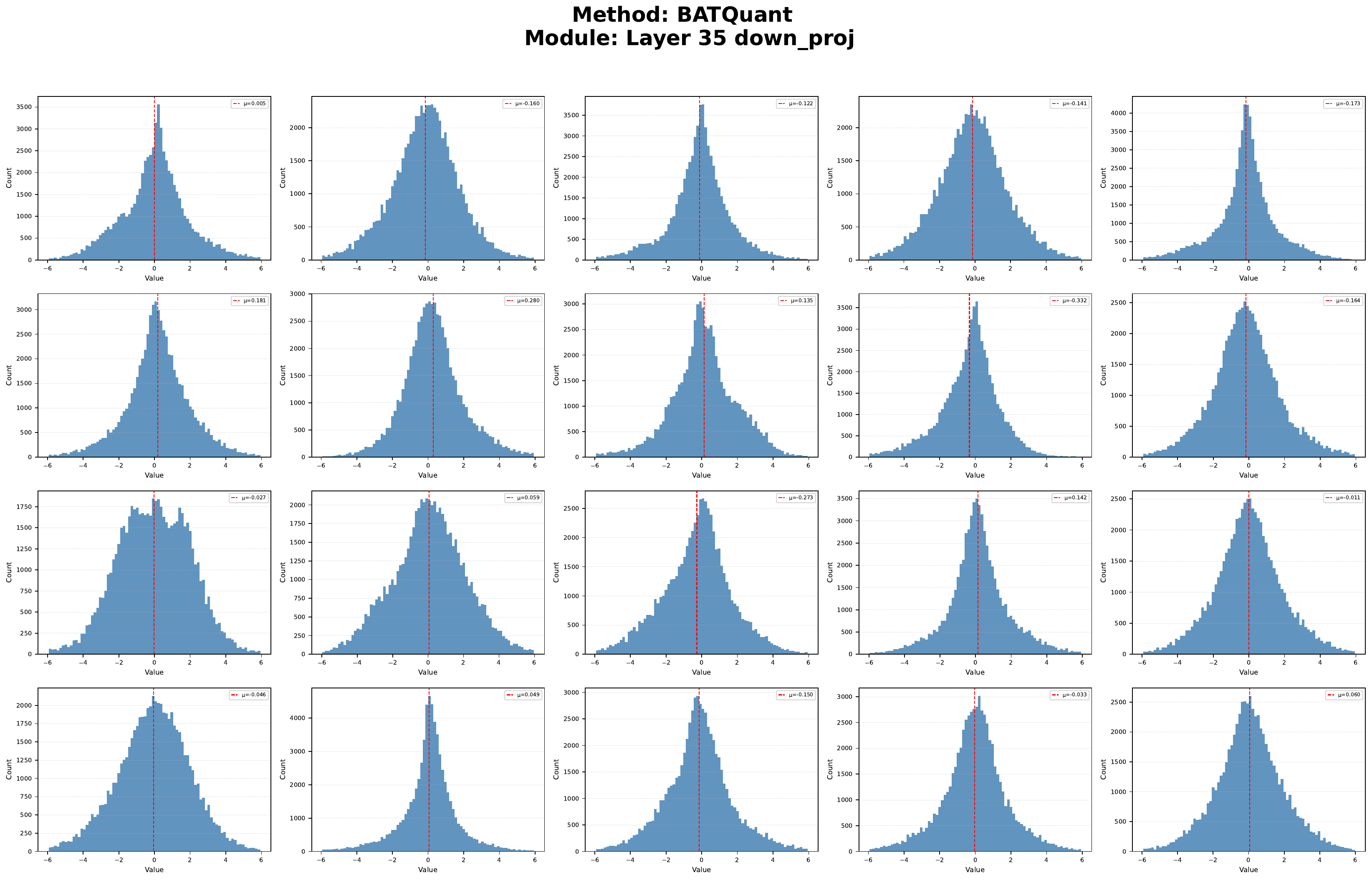}
    \caption{Activation distributions within different quantization blocks of the \texttt{down\_proj} module in layer 35 of Qwen3-8B with BATQuant (Ours).}
    \label{fig:activation_viz_batquant_all}
\end{figure}

Here, we provide the full details of activation distributions within different quantization blocks as shown in Figure~\ref{fig:activation_viz_rtn_all}, Figure~\ref{fig:activation_viz_brq_all}, Figure~\ref{fig:activation_viz_quarot_all} and Figure~\ref{fig:activation_viz_batquant_all}.

\subsection{Case Studies}
\begin{figure}[!t]
\centering
\begin{promptbox}
    \textbf{Prompt:}
    
    \vspace{0.5em}
    \begin{center}
        \includegraphics[width=0.4\linewidth]{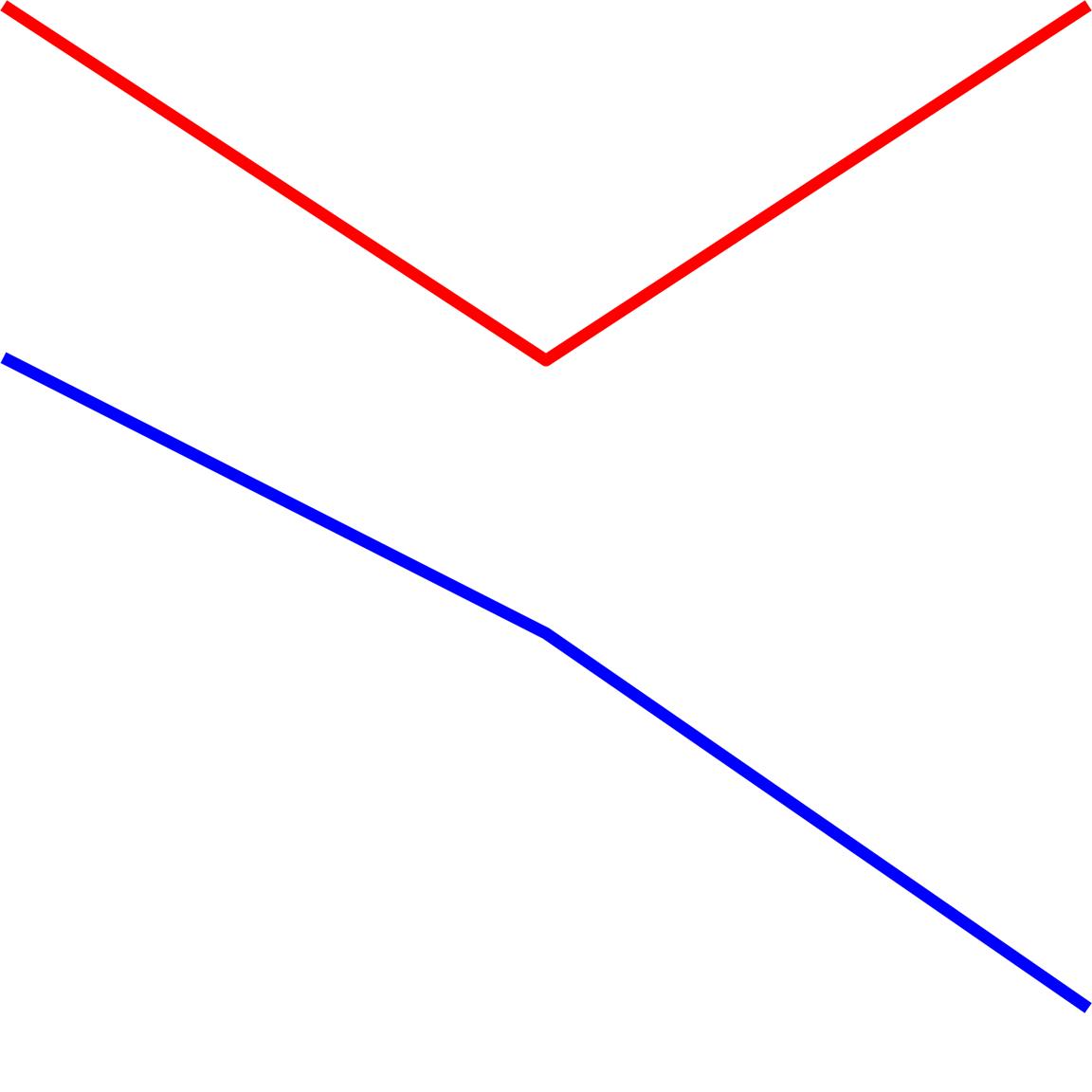} 
    \end{center}
    \vspace{0.5em}
    % ----------------------

    \noindent\begin{lstlisting}[
        language=Python,
        basicstyle=\small\ttfamily\color{black},
        keywordstyle=\color{black},
        commentstyle=\color{black},
        stringstyle=\color{black},
        showstringspaces=false,
        breaklines=true,
        numbers=none,
        xleftmargin=0pt,
        xrightmargin=0pt,
        breakindent=0pt, 
        aboveskip=0pt,
        belowskip=0pt
    ]
Question: Count the intersection points where the blue and red lines meet. Put your answer in curly brackets, e.g., {2}.
    \end{lstlisting}

    \vspace{0.5em}
    \textbf{Response:} 
    
    % --- 基线模型 (RTN/BRQ) - 错误示例 ---
    \begin{baselinebox}
        \raggedright
        % BF16 正确回答 (作为参考)
        \noindent\begin{lstlisting}[
            language=python,
            basicstyle=\footnotesize\ttfamily\color{darkgreen!80},
            keywordstyle=\color{darkgreen!80},
            commentstyle=\color{darkgreen!80},
            stringstyle=\color{darkgreen!80},
            literate={(model)}{{\color{darkgreen!80}\bfseries\normalsize\rmfamily{BF16}}}{4},
            showstringspaces=false,
            breaklines=true,
            numbers=none,
            xleftmargin=0pt,
            xrightmargin=0pt,
            breakindent=0pt, 
            aboveskip=0pt,
            belowskip=0pt
        ]
(model)
{0}
        \end{lstlisting}
        
        \noindent\rule{\linewidth}{0.4pt}
        \vspace{0.2em}
        
        % BRQ (W4A4) 错误回答
        \noindent\begin{lstlisting}[
            language=Python,
            basicstyle=\footnotesize\ttfamily\color{bad},
            keywordstyle=\color{bad},
            commentstyle=\color{bad},
            stringstyle=\color{bad},
            literate={(model)}{{\color{bad}\bfseries\normalsize\rmfamily{BRQ (W4A4KV16)}}}{17},
            showstringspaces=false,
            breaklines=true,
            numbers=none,
            xleftmargin=0pt,
            % xrightmargin=0pt,
            breakindent=0pt, 
            aboveskip=0pt,
            belowskip=0pt
        ]
(model)
There is one intersection point where the blue and red lines meet. The red lines form a "V" shape, and the blue line runs diagonally across the image. They intersect at a single point. 

{1}
        \end{lstlisting}
    \end{baselinebox}

    % --- BATQuant 模型 - 正确示例 ---
    \begin{freequantbox}
        \raggedright
        \noindent\begin{lstlisting}[
            language=Python,
            basicstyle=\footnotesize\ttfamily\color{darkgreen},
            keywordstyle=\color{darkgreen},
            commentstyle=\color{darkgreen},
            stringstyle=\color{darkgreen},
            literate={(model)}{{\color{darkgreen}\bfseries\normalsize\rmfamily{BATQuant (W4A4KV16)}}}{23},
            showstringspaces=false,
            breaklines=true,
            numbers=none,
            xleftmargin=0pt,
            xrightmargin=0pt,
            breakindent=0pt, 
            aboveskip=0pt,
            belowskip=0pt
        ]
(model)
{0}
        \end{lstlisting}
    \end{freequantbox}
    
\end{promptbox}

\caption{\textbf{Case study of {Qwen3-VL-8B-Instruct} on VLMBlind}. The input includes a real image (shown above) and a text question asking to count intersection points. Compared with the BRQ method which fails by hallucinating an intersection ({1}), BATQuant correctly identifies that there are no intersections ({0}), matching the BF16 baseline.}
\label{fig:case_qwenvl_vlmblind}
\end{figure}

\begin{figure}[!t]
\centering
\begin{promptbox}
    \textbf{Prompt:}
    
    % --- 图片插入区域 ---
    \vspace{0.5em}
    \begin{center}
        \includegraphics[width=0.6\linewidth]{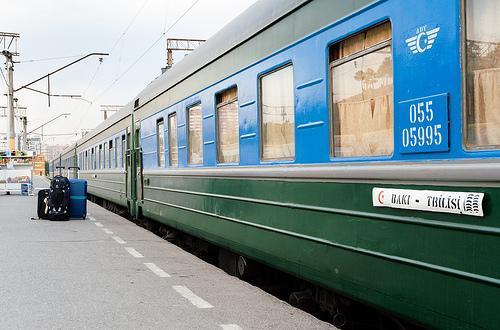} 
    \end{center}
    \vspace{0.5em}
    % ----------------------

    % --- 文本部分 (Question) ---
    \noindent\begin{lstlisting}[
        language=Python,
        basicstyle=\small\ttfamily\color{black},
        keywordstyle=\color{black},
        commentstyle=\color{black},
        stringstyle=\color{black},
        showstringspaces=false,
        breaklines=true,
        numbers=none,
        xleftmargin=0pt,
        xrightmargin=0pt,
        breakindent=0pt, 
        aboveskip=0pt,
        belowskip=0pt
    ]
What is the train number?
    \end{lstlisting}

    \vspace{0.5em}
    \textbf{Response:} 
    
    % --- 基线模型 (RTN/BRQ) - 错误示例 ---
    \begin{baselinebox}
        \raggedright
        % BF16 正确回答 (作为参考)
        \noindent\begin{lstlisting}[
            language=python,
            basicstyle=\footnotesize\ttfamily\color{darkgreen!80},
            keywordstyle=\color{darkgreen!80},
            commentstyle=\color{darkgreen!80},
            stringstyle=\color{darkgreen!80},
            literate={(model)}{{\color{darkgreen!80}\bfseries\normalsize\rmfamily{BF16}}}{4},
            showstringspaces=false,
            breaklines=true,
            numbers=none,
            xleftmargin=0pt,
            xrightmargin=0pt,
            breakindent=0pt, 
            aboveskip=0pt,
            belowskip=0pt
        ]
(model)
055 05995
        \end{lstlisting}
        
        \noindent\rule{\linewidth}{0.4pt}
        \vspace{0.2em}
        
        % BRQ (W4A4) 错误回答
        \noindent\begin{lstlisting}[
            language=Python,
            basicstyle=\scriptsize\ttfamily\color{bad},
            keywordstyle=\color{bad},
            commentstyle=\color{bad},
            stringstyle=\color{bad},
            literate={(model)}{{\color{bad}\bfseries\normalsize\rmfamily{BRQ (W4A4KV16)}}}{21},
            showstringspaces=false,
            breaklines=true,
            numbers=none,
            xleftmargin=0pt,
            % xrightmargin=0pt,
            breakindent=0pt, 
            aboveskip=0pt,
            belowskip=0pt
        ]
(model)
Based on the image provided, the train number is **055**.

This number is clearly visible on the side of the train car in the image. It appears as a white or light-colored number (likely "055") 

...

The train appears to be a passenger train, likely operating within Vietnam given the Vietnamese language used in the markings.
        \end{lstlisting}
    \end{baselinebox}

    % --- BATQuant 模型 - 正确示例 ---
    \begin{freequantbox}
        \raggedright
        \noindent\begin{lstlisting}[
            language=Python,
            basicstyle=\scriptsize\ttfamily\color{darkgreen},
            keywordstyle=\color{darkgreen},
            commentstyle=\color{darkgreen},
            stringstyle=\color{darkgreen},
            literate={(model)}{{\color{darkgreen}\bfseries\normalsize\rmfamily{BATQuant (W4A4KV16)}}}{28},
            showstringspaces=false,
            breaklines=true,
            numbers=none,
            xleftmargin=0pt,
            xrightmargin=0pt,
            breakindent=0pt, 
            aboveskip=0pt,
            belowskip=0pt
        ]
(model)
Based on the image provided, the train number is **055 05995**.

This number is clearly visible on the side of the train car, just below the window. The "055" appears to be the train's route or service number, and "05995" is likely its specific car or vehicle number within that service.
        \end{lstlisting}
    \end{freequantbox}
    
\end{promptbox}

\caption{\textbf{Case study of {Qwen3-VL-8B-Instruct} on OCRBench}. The input includes a real image of a train and a question asking for the train number. Compared with the BRQ method which fails by only recognizing partial information ("055"), BATQuant correctly identifies the full train number ("055 05995"), matching the BF16 baseline.}
\label{fig:case_qwenvl_ocrbench}
\end{figure}

We qualitatively compare BATQuant against {BRQ} (W4A4) on geometric reasoning and OCR tasks under \texttt{W4A4KV16} senario. As shown in Figures~\ref{fig:case_qwenvl_vlmblind} and~\ref{fig:case_qwenvl_ocrbench}, while BRQ suffers from feature distortion leading to hallucinations, BATQuant preserves critical visual details matching the BF16 baseline. In Figure~\ref{fig:case_qwenvl_vlmblind}, the task requires counting line intersections. The BRQ baseline incorrectly hallucinates an intersection point (\texttt{\{1\}}), likely due to quantization noise distorting edge continuity. In contrast, BATQuant correctly identifies zero intersections (\texttt{\{0\}}), demonstrating superior preservation of spatial structures. Figure~\ref{fig:case_qwenvl_ocrbench} presents a challenging train number recognition task. BRQ fails to capture the full sequence, truncating the answer to ``055''. Conversely, BATQuant accurately recovers the complete number ``055 05995'', proving its effectiveness in retaining high-frequency details essential for dense text recognition. These cases highlight that unlike BRQ, which struggles with subtle visual cues under aggressive quantization, BATQuant robustly maintains semantic fidelity.

\end{document}